\documentclass[12pt]{article}
\usepackage{amsmath}
\usepackage{times}
\usepackage{graphicx}
\usepackage{color}
\usepackage{multirow}
\usepackage[natbibapa]{apacite}
\usepackage{rotating}
\usepackage{bbm}
\usepackage{latexsym}
\usepackage{main}		
\usepackage[hidelinks]{hyperref}		
\usepackage{url}		
\usepackage{times}		
\usepackage{amssymb}		
\usepackage{amsmath}		
\usepackage{amsthm}		
\usepackage{graphicx}		
\usepackage{multicol}		
\usepackage{booktabs}		
\usepackage{resizegather}		
\usepackage{caption}		
\usepackage{color}		
\usepackage{bm}		
\usepackage{algorithm}
\usepackage[noend]{algpseudocode}
\usepackage{diagbox}		
\usepackage{subcaption}		

\newcommand{\captionfonts}{\normalsize}

\makeatletter  
\long\def\@makecaption#1#2{%
  \vskip\abovecaptionskip
  \sbox\@tempboxa{{\captionfonts #1: #2}}%
  \ifdim \wd\@tempboxa >\hsize
    {\captionfonts #1: #2\par}
  \else
    \hbox to\hsize{\hfil\box\@tempboxa\hfil}%
  \fi
  \vskip\belowcaptionskip}
\makeatother   
\begin{document}
\hspace{13.9cm}1

\ \vspace{20mm}\\

\newcommand{\xaxis}{\textit{x-axis }}	\newcommand{\yaxis}{\textit{y-axis }}		
\newcommand{\NN}{\textit{Net2Net}}		
\def\mW{{\bm{W}}}		\def\sR{{\mathbb{R}}}		
\def\mU{{\bm{U}}}

{\LARGE Towards Training Recurrent Neural Networks \\ for Lifelong Learning}

% \ \\
% {\bf \large Shagun Sodhani*$^{\displaystyle 1}$, Sarath Chandar*$^{\displaystyle 1}$, Yoshua Bengio$^{\displaystyle 1}$}\\
% {$^{\displaystyle 1}$Mila, University of Montreal, {\small \tt name.lastname@umontreal.ca}}\\
% {$^{\displaystyle *}$Equal Contribution\\
% %

\ \\
{\bf \large Shagun Sodhani$^{\displaystyle 1*}$, Sarath Chandar$^{\displaystyle 1*}$, Yoshua Bengio$^{\displaystyle 1, 2}$}\\
{$^{\displaystyle 1}$Mila, Universit\'{e} de Montr\'{e}al, Canada} \\
{$^{\displaystyle 2}$CIFAR Senior Fellow}\\
{$^{\displaystyle *}$Equal Contribution\\
{{\small \tt Contact: sshagunsodhani@gmail.com, apsarathchandar@gmail.com}}\\

%\ \\[-2mm]

\thispagestyle{empty}
\markboth{}{NC instructions}
\ \vspace{-0mm}\\

\begin{center} {\bf Abstract} \end{center}

Catastrophic forgetting and capacity saturation are the central challenges of any parametric lifelong learning system. In this work, we study these challenges in the context of sequential supervised learning with an emphasis on recurrent neural networks. To evaluate the models in the lifelong learning setting, we propose a curriculum-based, simple, and intuitive benchmark where the models are trained on tasks with increasing levels of difficulty. To measure the impact of catastrophic forgetting, the model is tested on all the previous tasks as it completes any task. As a step towards developing \textit{true} lifelong learning systems, we unify \textit{Gradient Episodic Memory} (a catastrophic forgetting alleviation approach) and \textit{Net2Net} (a capacity expansion approach). Both these models are proposed in the context of feedforward networks and we evaluate the feasibility of using them for recurrent networks. Evaluation on the proposed benchmark shows that the unified model is more suitable than the constituent models for lifelong learning setting.

\section{Introduction}
\label{sec::introduction}

Lifelong Machine Learning considers systems that can learn many tasks (from one or more domains) over a lifetime \citep{thrun1998lifelong,silver2013lifelong}. This has several names and manifestations in the literature: incremental learning \citep{solomonoff1989system}, continual learning \citep{ring1997child}, explanation-based learning \citep{thrun1996explanation,thrun2012explanation}, never-ending learning \citep{carlson2010toward}, etc. The underlying idea motivating these efforts is the following: Lifelong learning systems would be more effective at learning and retaining knowledge across different tasks. By using the prior knowledge and exploiting similarity across tasks, they would be able to obtain better priors for the task at hand. Lifelong learning techniques are very important for training intelligent autonomous agents that would need to operate and make decisions over extended periods of time. These characteristics are especially important in the industrial setups where the deployed machine learning models are being updated frequently with new incoming data whose distribution need not match the data on which the model was originally trained. 

The lifelong learning paradigm is not just restricted to the multi-task setting with clear task boundaries. In real life, the system has no control over what task it receives at any given time step. In such situations, there is no clear task boundary. Lifelong learning is also relevant when the system is learning just a single task but the data distribution changes over time.

Lifelong learning is an extremely challenging task for machine learning models because of two primary reasons:

\begin{enumerate}
    \item \textbf{Catastrophic Forgetting}: As the model is trained on a new task (or a new data distribution), it is likely to forget the knowledge it acquired from the previous tasks (or data distributions). This phenomenon is also known as the \textit{catastrophic interference} \citep{mccloskey1989catastrophic}.
    \item \textbf{Capacity Saturation}: Any parametric model, however large, can only have a fixed amount of representational capacity, to begin with. Given that we want the model to retain knowledge as it progresses through multiple tasks, the model would eventually run out of capacity to store the knowledge acquired in the successive tasks. The only way for it to continue learning, while retaining previous knowledge, is to increase its capacity on the fly.
\end{enumerate}

Catastrophic forgetting and capacity saturation are related issues. In fact, capacity saturation can lead to catastrophic forgetting. But it is not the only cause for catastrophic forgetting. As the model is trained on one data distribution for a long time, it can forget its ``learning'' from the previous data distributions irrespective of how much effective capacity it has. While an under-capacity model could be more susceptible to catastrophic forgetting, having sufficient capacity (by say using very large models) does not protect against catastrophic forgetting (as we demonstrate in section \ref{sec::experiment}). Interestingly, a model that is immune to catastrophic forgetting could be more susceptible to capacity saturation (as it uses more of its capacity to retain the previously acquired knowledge). We demonstrate this effect as well in section \ref{sec::experiment}. It is important to think about both catastrophic forgetting and capacity saturation together as solving just one problem does not take care of the other problem. Further, the role of capacity saturation and capacity expansion in lifelong learning is an under-explored topic.

Motivated by these challenges, we compile a list of desirable properties that a model should fulfill to be deemed suitable for lifelong learning settings:

\begin{enumerate}
    \item \textbf{Knowledge Retention} - As the model learns to solve new tasks, it should not forget how to solve the previous tasks.
    \item \textbf{Knowledge Transfer} - The model should be able to reuse the knowledge acquired during previous tasks to solve the current task. If the tasks are related, this knowledge transfer would lead to faster learning and better generalization over the lifetime of the model.
    \item \textbf{Parameter Efficiency} - The number of parameters in the model should ideally be bounded, or grow at-most sub-linearly as new tasks are added.
    \item \textbf{Model Expansion} - The model should be able to increase its capacity on the fly by ``expanding" itself.
\end{enumerate}

The model expansion characteristic comes with additional constraints: In a true lifelong learning setting, the model would experience a continual stream of training data that can not be stored. Hence any model would, at best, have access to only a small sample of the historical data. In such a setting, we can not rely on past examples to train the expanded model from scratch and a zero-shot knowledge transfer is desired. Considering the parameter efficiency and the model expansion qualities together implies that we would also want the computational and memory costs of the model to increase only sublinearily as the model trains on new tasks.

We propose to unify the Gradient Episodic Memory (GEM) model \citep{NIPS2017_7225} and the \NN~framework \citep{chen2015net2net} to develop a model suitable for lifelong learning. The GEM model provides a mechanism to alleviate catastrophic forgetting, while allowing for improvement in the previous tasks by beneficial backward transfer of knowledge. \NN~is a technique for transferring knowledge from a smaller, trained neural network to another larger, untrained neural network. We discuss both these models in detail in the Related Work (section \ref{sec::related-work}). 

One reason hindering research in lifelong learning is the absence of standardized training and evaluation benchmarks. For instance, the vision community benefited immensely from the availability of the Imagenet dataset \citep{imagenet} and we believe that availability of a standardized benchmark would help to propel and streamline research in the domain of lifelong learning. Creating a good benchmark set up to study different aspects of lifelong learning is extremely challenging. \citet{pmlr-v78-lomonaco17a} proposed a new benchmark for Continuous Object Recognition (CORe50) in the context of computer vision. \citet{NIPS2017_7225} considered different variants of MNIST and CIFAR-100 datasets for lifelong supervised learning. These benchmarks help study-specific challenges like catastrophic forgetting by abstracting out the other challenges but they are quite far from real-life setting. Another limitation of the existing benchmarks is that they are largely focused on non-sequential tasks and there has been no such benchmark available for lifelong learning in the context of sequential supervised learning. Sequential supervised learning, like reinforcement learning, is a sequential task and hence more challenging than one step supervised learning tasks. However, unlike reinforcement learning, the setup is still supervised and hence is easier to focus on the challenges in lifelong learning in isolation from the challenges in reinforcement learning.

In this work, we propose a curriculum-based, simple and intuitive benchmark for evaluating lifelong learning models in the context of sequential supervised learning. We consider a single task setting where the model starts with the first data distribution (the simplest data distribution) and subsequently progresses to the more difficult data distributions. We can consider each data distribution as a task by itself. Each task has well-defined criteria of completion and the model can start training on a task only after learning over all the previous tasks in the curriculum. Each time the model finishes a task, it is evaluated on all the tasks in the curriculum (including the tasks that it has not been trained on so far) so as to compare the performance of the model in terms of both catastrophic forgetting (for the previously seen tasks) and generalization (to unseen tasks). 

If the model fails to learn a task (as per pre-defined criteria of success), we expand the model and let it train on the current task again. The expanded model is again evaluated on all the tasks just like the regular, unexpanded model. Performing this evaluation step enables us to analyze and understand how the model expansion step affects the model's capabilities in terms of generalization and catastrophic forgetting. We describe the benchmark and the different tasks in detail in the Tasks and Setup (section \ref{sec::task}).

Our main contributions are as follows:
\begin{enumerate}
    \item We tackle the two main challenges of lifelong learning by unifying Gradient Episodic Memory (a lifelong learning technique to alleviate catastrophic forgetting) with \NN~(a capacity expansion technique).
    \item We propose a simple benchmark of tasks for training and evaluating models for learning sequential problems in the lifelong learning setting.
    \item We show that both GEM and \NN~which are originally proposed for feed-forward architectures are indeed useful for recurrent neural networks as well.
    \item We evaluate the proposed unified model on the proposed benchmark and show that the unified model is better suited to the lifelong learning setting as compared to the two constituent models.
\end{enumerate}

\section{Related Work}
\label{sec::related-work}
We review the prominent works dealing with catastrophic forgetting,  capacity saturation and model expansion as these are the important aspects of lifelong learning.  

\subsection{Catastrophic Forgetting}

Much of the work in the domain of catastrophic forgetting can be broadly classified into two approaches:

\begin{enumerate}
    \item \textbf{Model Regularization}: A common and useful strategy is to freeze parts of the model as it trains on successive tasks. This can be seen as locking in the knowledge about how to solve different tasks in different parts of the model so that training on the subsequent tasks can not interfere with this knowledge. Sometimes, the weights are not completely frozen and are regularized to not change \textit{too-much} as the model train across different tasks. This approach is adopted by elastic weight consolidation (EWC) model \citep{2016arXiv161200796K}. As the model train through the sequence of tasks, the learning is slowed down for weights which are important to the previous tasks. \citet{liu2018rotate} extended this model by reparameterizing the network to approximately diagonalize the Fisher information matrix of the network parameters. This reparameterization leads to a factorized rotation of the parameter space and makes the diagonal Fisher Information Matrix assumption (of the EWC model) more applicable. \cite{reimannianwalk} presented RWalk, a generalization of EWC and Path
Integral~\citep{pathintegral} with a theoretically grounded KL-divergence based perspective along with several new metrics. One downside of such approaches is the loss in the effective trainable capacity of the model as more and more model parameters are regularized over time. This seems counter-intuitive given the desirable properties that we want the lifelong learning systems to have (section \ref{sec::introduction}).
    
    \item \textbf{Rehearsing using previous examples}: When learning on a given task, the model is also shown examples from the previous tasks. This \textit{rehearsal} setup \citep{silver2002task} can help in two ways - if the tasks are related, training on multiple tasks helps in transferring knowledge across the tasks. If the tasks are unrelated, the setup still helps to protect against catastrophic forgetting. \cite{rebuffi2017icarl} proposed the \textit{iCaRL} model which focuses on the class-incremental learning setting whereas the number of classes (in the classification system) increase, the model is shown examples from the previous tasks. Generally, this strategy requires persisting some training examples per task. In practice, the cost of persisting some data samples (in terms of memory requirements) is much smaller than the memory requirements of the model. Though, in the rehearsal setup, the computational cost of training the model increases with each new task as the model has to \textit{rehearse} on the previous tasks as well. 
    
\end{enumerate}

\cite{mensink2012metric} proposed the \textit{Nearest Mean Classifier} (NCM) model in the context of large scale, multi-class image classification. The idea is to use distance-based classifiers where a training example is assigned to the class which is ``nearest" to it. The setup allows adding new classes and new training examples to existing classes at a near-zero cost. Thus the system can be updated on the fly as more training data becomes available. Further, the model could periodically be trained on the complete dataset (collected thus far). \cite{2016arXiv160609282L} proposed the \textit{Learning without Forgetting}(LwF) approach in the context of computer vision tasks. The idea is to divide the model into different components. Some of these components are shared between different tasks and some of the components are task-specific. When a new task is introduced, first the existing network is used to make predictions for the data corresponding to the new task. These predictions are used as the ``ground-truth'' labels to compute a regularization loss that ensures that training on the new task does not affect model's performance on the previous task. Then a new task-specific component is added to the network and the network is trained to minimize the sum of loss on the current task and the regularisation loss. The ``addition'' of new components per task makes the LwF model parameter inefficient.

\citet{2016arXiv160609282L} proposed to use the distillation principle \citep{hinton2015distilling} to incrementally train a single network for learning multiple tasks by using data only from the current task. \citet{NIPS2017_7051} proposed incremental moment matching (IMM) which incrementally matches the moment of the posterior distribution of the neural network which is trained on the first and the second task, respectively. While this approach seems to give strong results, it is evaluated only on datasets with very few tasks. \citet{hat} proposed to use hard attention targets (HAT) to learn pathways in a given base network using the id of the given task. The pathways are used to obtain the task-specific networks. The limitation of this approach is that it requires knowledge about the current task id.

The recently proposed Gradient Episodic Memory approach \citep{NIPS2017_7225} outperforms many of these models while enabling positive transfer on the backward tasks. It uses an episodic memory which stores a subset of the observed examples from each task. When training on a given task, an additional constraint is added such that the loss on the data corresponding to the previous tasks does not increase though it may or may not decrease. One limitation of the model is the need to compute gradients corresponding to the previous task at each learning iteration. Given that GEM needs to store only a few examples per task (in our experiments, we stored just one batch of examples), the storage cost is negligible. Given the strong performance and low memory cost, we use GEM as the first component of our unified model.

\subsection{Capacity Saturation and Model Expansion}

The problem of capacity saturation and model expansion has been extensively studied from different perspectives. Some works explored model expansion as a means of transferring knowledge from a small network to a large network to ease the training of deep neural networks \citep{gutstein2008knowledge,2018arXiv180504770F}. Analogously, the idea of distilling knowledge from a larger network to a smaller network has been explored in \citep{hinton2015distilling,romero2014fitnets} etc. Majority of these approaches focus on training the new network on a single supervised task where the data distribution does not change much and the previous examples can be reused several times. This is not possible in a true online lifelong learning setting where the model experiences a continual stream of training data and has no access to previously seen examples again.

\citet{chen2015net2net} proposed using function-preserving transformations to expand a small, trained network (referred to as the teacher network) into a large, untrained network (referred to as the student network). Their primary motivation was to accelerate the training of large neural networks by first training small neural networks (which are easier and faster to train) and then transferring their knowledge to larger neural networks. The paper evaluated the technique in the context of single task supervised learning and mentioned continual learning as one of the motivations. Given that \textit{Net2Net} enables the zero-shot transfer of knowledge to the expanded network, we use this idea of function preserving transformations to achieve zero-shot knowledge transfer in the proposed unified model.

\begin{table}[htbp]
\scriptsize
\centering
\caption{Comparison of different models in terms of the desirable properties they fulfill.}
\label{tab::model-comp}
\resizebox{\columnwidth}{!}{%
\begin{tabular}{|c|c|c|c|c|}
\hline
 \diagbox[width=6em]{Model}{Property}                                                               & \begin{tabular}[c]{@{}c@{}}Knowledge\\  Retention\end{tabular} & \begin{tabular}[c]{@{}c@{}}Knowledge \\ Transfer\end{tabular} & \begin{tabular}[c]{@{}c@{}}Parameter \\ Efficiency\end{tabular} & \begin{tabular}[c]{@{}c@{}}Model \\ Expansion\end{tabular} \\ \hline

EWC                 & \checkmark    &                             & \checkmark     &                           \\ \hline
IMM                    & \checkmark    & \checkmark   & \checkmark                              &                           \\ \hline
iCaRL            & \checkmark    &                             &      \checkmark                         &                           \\ \hline
NCM                  & \checkmark    &                             &   \checkmark                            &                           \\ \hline
LwF               & \checkmark    &                             &                               &                           \\ \hline
GEM       & \checkmark    &    \checkmark                         &    \checkmark                           &                           \\ \hline
Net2Net          &                              &    & \checkmark     & \checkmark \\ \hline
Progressive Nets    & \checkmark    & \checkmark   &                             & \checkmark 
\\ \hline
Network of Experts & \checkmark    & \checkmark   &                               & \checkmark \\ \hline
Piggyback & \checkmark    &    &                               & \checkmark \\ \hline
HAT & \checkmark    &  \checkmark  &            \checkmark                   &  \\ \hline
\end{tabular}
}
\end{table}

\citet{rusu2016progressive} proposed the idea of Progressive Networks that explicitly supports the transfer of features across a sequence of tasks. The progressive network starts with a single \textit{column} or model (neural network) and new \textit{columns} are added as more tasks are encountered. Each time the network learned a task, the newly added \textit{column} (corresponding to the task) is ``frozen" to ensure that ``knowledge" can not be lost. Each new \textit{column} uses the layer-wise output from all the previous columns to explicitly enable transfer learning. As a new \textit{column} is added per task, the number of \textit{columns} (and hence the number of network parameters) increases linearly with the number of tasks. Further, when a new \textit{column} is added, only a fraction of the new capacity is actually utilized, thus each new column is increasingly underutilized. Another limitation is that during training, the model explicitly needs to know when a new task starts so that a new \textit{column} can be added to the network. Similarly, during inference, the network needs to know the task to which the current data point belongs to so that it knows which \textit{column} to use. \citet{2016arXiv161106194A} build upon this idea and use a Network of Experts where each expert model is trained for one task. During inference, a set of gating autoencoders are used to select the expert model to query. This gating mechanism helps to reduce the dependence on knowing the task label for the test data points.

\citet{mallya2018piggyback} proposed the piggyback approach to train the model on a base task and then learn different bit masks (for parameters in the base network) for different tasks. One advantage as compared to Progressive Networks is that only 1 bit is added per parameter of the base model (as compared to 1 new parameter per parameter of the base model). The shortcoming of the approach, however, is that knowledge can be transferred only from the base task to the subsequent tasks and not between different subsequent tasks.

Table \ref{tab::model-comp} compares the different lifelong learning models in terms of the desirable properties they fulfill. The table makes it very easy to determine which combination of models could be feasible. If we choose a parameter-inefficient model, then the unified model will be parameter inefficient which is clearly undesirable. Further, we want at least one of the component models to have expansion property so that the capacity can be increased on the fly. This narrows down the choice of the first model to \NN. Since this model lacks both knowledge retention and knowledge transfer, we could pick either IMM, GEM or HAT as the second component. IMM is evaluated for very few tasks while HAT requires the task ids to be known beforehand. In contrast, GEM is reported to work well for a large number of tasks \citep{NIPS2017_7225}. Given these considerations, we choose GEM as the second component. Now, the unified model has all the four properties.

\section{Tasks and Benchmark}
\label{sec::task}
In this section, we describe the tasks, training, and the evaluation setup that we proposed for benchmarking the lifelong learning models in the context of sequential supervised learning. In a true lifelong learning setting, the training distribution can change arbitrarily and no explicit demarcation exists between the data distribution corresponding to the different tasks. This makes it extremely hard to study how model properties like catastrophic forgetting and generalization capability evolve with the training. We sidestep these challenges by using a curriculum-based, simple and intuitive setup where we can have full control over the training data distributions. This setup gives us explicit control over when the model experiences different data distributions and in what order. Specifically, we train the models in the curriculum style setup \citep{bengio2009curriculum} where the tasks are ordered by difficulty. We discuss the rationale behind using the curriculum approach in section \ref{sec::task::discussion::rationale}. We consider the following three tasks as part of the benchmark:

\subsection{Copy Task}
The copy task is an algorithmic task introduced in \citep{2014arXiv1410.5401G} to test whether the training network can learn to store and recall a long sequence of random vectors. Specifically, the network is presented with a sequence of randomly initialized, seven-bit vectors. Each such vector is followed by an eighth bit which serves as a delimiter flag. This flag is zero at all time steps except for the end of the sequence. The network is trained to generate the entire sequence except the delimiter flag. The different levels are defined by considering input sequences of different lengths. We start with input sequences of length 5 and increase the sequence length in the steps of 3 and go till the maximum sequence length of 62 (20 levels). We can consider arbitrarily large sequences but we restrict ourselves to maximum sequence length of 62 as none of the considered models were able to learn all these sequences. We report the bit-wise accuracy metric.

\subsection{Associative Recall Task}
The associative recall task is another algorithmic task introduced in \citep{2014arXiv1410.5401G}. In this task, the network is shown a list of items where each item is a sequence of randomly initialized 8-bit binary vectors, bounded on the left and the right by the delimiter symbols. First, the network is shown a sequence of items and then it is shown one of the items (from the sequence). The model is required to output the item that appears next from the ingested sequence. We set the length of each item to be 3. The levels are defined in terms of the number of items in the sequence. The first level considers sequences with 5 items and the number of items is increased in steps of 3 per level, going till 20 levels where there are 62 items per sequence.  We report the bit-wise accuracy metric.

\subsection{Sequential Stroke MNIST Task}
Sequential Stroke MNIST (SSMNIT) task was introduced in \citep{DBLP:journals/corr/GulcehreCB17} with an emphasis on testing the long-term dependency modeling capabilities of the RNNs. In this task, each MNIST digit image \textit{I} is represented as a sequence of quadruples 
$\{dx_i, , dy_i, eos_i, eod_i\}_{i=1}^{T}$. Here, $T$ is the number of pen strokes needed to define the digit, $(dx_i, dy_i)$ denotes the pen offset from the previous to the current stroke (can be 1, -1 or 0), $eos_i$ is a binary-valued feature to denote end of stroke and $eod_i$ is another binary-valued feature to denote end of the digit. The average number of strokes per digit is 40. Given a sequence of pen-stroke sequences, the task is to predict the sequence of digits corresponding to each pen-stroke sequences in the given order. This is an extremely challenging task as the model is first required to predict the digits based on the pen-stroke sequence, count the number of digits, and then generate the digits in the same order as the input after having processed the entire sequence of pen-strokes. The levels are defined in terms of the number of digits that make up the sequence.  Given that this task is more challenging than the other two tasks, we use a sequence of length 1 (i.e. single digit sequences) for the first level and increase the sequence length in steps of 1. Just like before, we consider 20 levels and report the per-digit accuracy as the metric.

\subsection{Benchmark}
\label{sec::task::benchmark}
So far, we have considered the setup with three tasks and have defined multiple levels within each task. Alternatively, we could think of each task as a ``task distribution'' and each level (within the task) as a task (within a ``task distribution''). From now on, we employ the \textit{task-distribution / task} notation to keep the discussion consistent with the literature in lifelong learning where multiple tasks are considered. Thus we have 3 ``task distributions'' (Copy, Associative Recall, and SSMNIST) and multiple tasks (in increasing order of difficulty) per ``task distribution''. To be closely aligned with the \textit{true} lifelong learning setup, we train all the models in an online manner where the network sees a stream of training data. Further, none of the examples are seen more than once. A common setup in online learning is to train the model with one example at a time. Instead, we train the model using mini-batches of 10 examples at a time to exploit the computational benefits in using mini-batches. However, we ensure that every mini-batch is generated randomly and that none of the examples are repeated so that a separate validation or test dataset is not needed. For each task (within a ``task distribution''), we report the \textit{current task accuracy} as an indicator of the model's performance on the current task. If the running-average of the \textit{current task accuracy}, averaged over last $k$ batches, is greater-than or equal-to $c$\%, the model is said to have ``learned'' the task and we can start training the model on the next task. If the model fails to learn the current task, we stop the training procedure and report the number of tasks completed. Every model is trained for $m$ number of mini-batches before it is evaluated to check if it has ``learned'' the task. Since we consider models with different capacity, some models could ``learn'' the task faster thus experiencing fewer examples. This setup ensures that each model is trained on the same number of examples. This training procedure is repeated for all the ``task distributions''. $k$, $m$ and $c$ are the parameters of the benchmark and can be set to any reasonable value as long as they are kept constant for all tasks in a given ``task distribution''. Specifically, we set $k$ = 100, and $m$ = 10000 for all the tasks. $c$ = 80 for Copy and $c$ = 75 for Associative Recall and SSMNIST. 

In the lifelong learning setting, it is very important for the model to retain knowledge from the previous tasks while generalizing to the new tasks. Hence, each time the model ``learns'' a task, we evaluate it on all the previous tasks (that it has been trained on so far) and report the model's performance (in terms of accuracy) for each of the previous task. Additionally, we also report the average of all these previous task accuracies and denote it as the \textit{per-task-previous-accuracy}.  When the model fails to learn a task and its training is stopped, we report both the individual \textit{per-task-previous-accuracy} metrics and the average of these metrics, which is denoted as the \textit{previous-task-accuracy}. While \textit{per-task-previous-accuracy} metric can be used as a crude approximation to quantify the effect of catastrophic forgetting, we highlight that the metric, on its own, is an insufficient metric. Consider a model which learns to solve just 1 task and terminates training after the $2^{nd}$ task. When evaluated for backward transfer, it would be evaluated only on the $1^{st}$ task. Now consider a model which just finished training on the $10^{th}$ task. When evaluated for backward transfer, it would be evaluated on the first $9$ tasks. \textit{per-task-previous-accuracy} metric favors models which stop training early and hence the series of \textit{per-task-previous-accuracy} metrics is a more relevant measure.

Another interesting aspect of lifelong learning is the generalization to unseen tasks. Analogous to the \textit{per-task-previous-accuracy} and \textit{previous-task-accuracy}, we consider the \textit{per-task-future-accuracy} and \textit{future-task-accuracy}. There is no success criteria associated with this evaluation phase and the metrics are interpreted as a proxy of model's ability to generalize to future ``tasks''. In our benchmark, the tasks are closely related which makes it reasonable to test generalization to new tasks. Note that the benchmark tasks can have levels beyond 20 as well. We limited our evaluation to 20 levels as none of the models could complete all the levels.

In the context of lifelong learning systems, the model needs to expand its capacity once it has saturated to make sure it can keep learning from the incoming data. We simulate this scenario in our benchmark-setting as follows: If the model fails to complete a given task, we use some capacity expansion technique and expand the original model into a larger model. Specifically, since we are considering RNNs, we expand the size of the hidden state matrix. The expanded model is then allowed to train on the current task for 20000 iterations. From there, the expanded model is evaluated (and trained on subsequent tasks) just like a regular model. If the expanded model fails on any task, the training is terminated. Note that this termination criterion is a part of our evaluation protocol. In practice, we can evaluate the model as many times as we want. In the ablation studies, we consider a case where the model is expanded twice.

\begin{figure}
\centering
\includegraphics[scale=1.0]{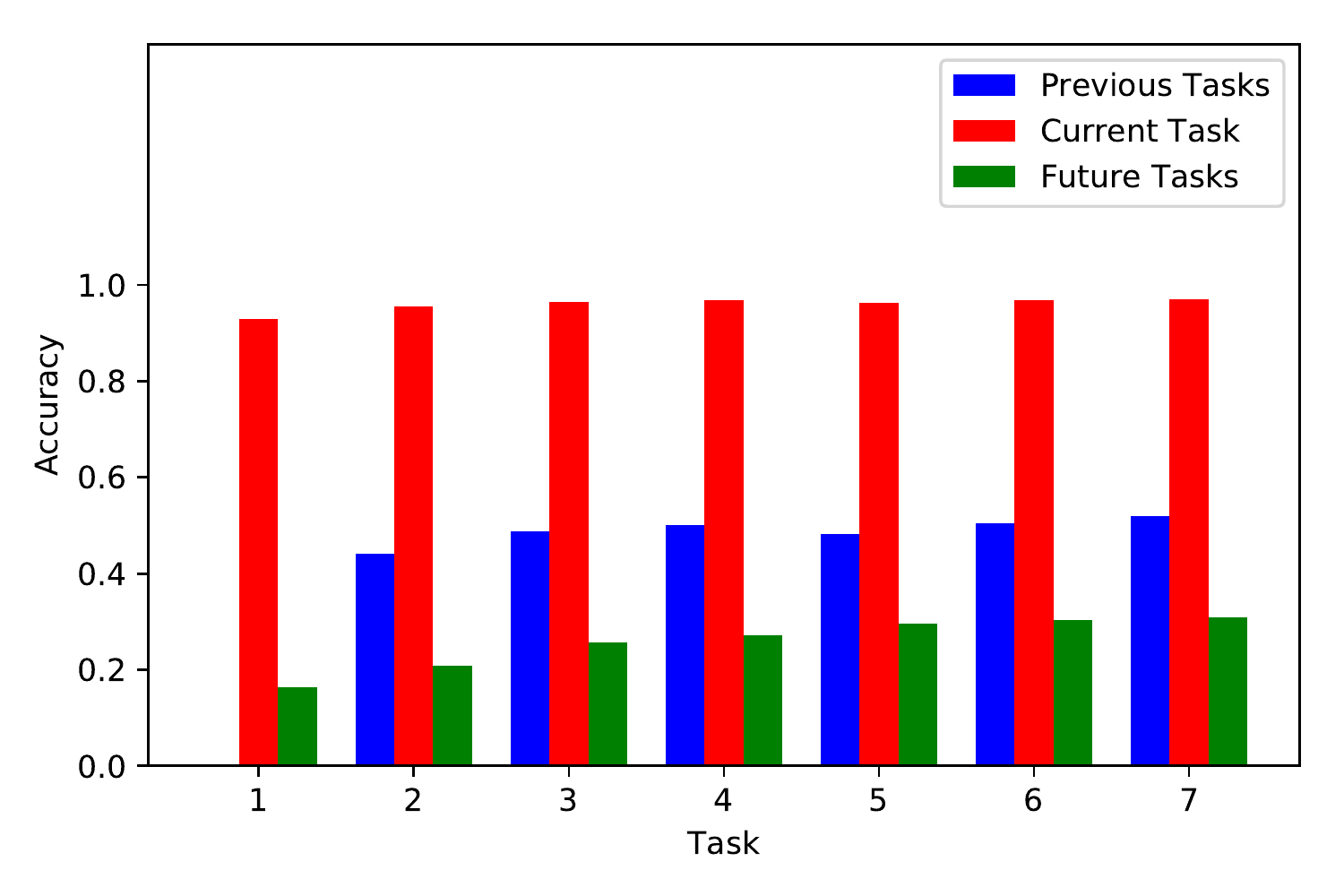}
\caption{Per-level accuracy on previous tasks, current task, and future tasks for a 128 dimensional LSTM trained in the SSMNIST ``task distribution''. The model heavily overfits to the sequence length.}
\label{fig:seq}
\end{figure}

\subsection{Rationale for using curriculum style setup}
\label{sec::task::discussion::rationale}
For all the three ``task distributions'', it can be reasonably argued that as the sequence length increases, the tasks become more challenging as the model needs to store/retrieve a much longer sequence. Hence, for each ``task distribution'', we define a curriculum of tasks by controlling the length of the input sequences. We note that our experimental setup is different from the real-life setting in two ways: First, in the real-life, we may not know beforehand as to which data point belongs to which data (or task) distribution. Second, in real life, we have no control over the difficulty or complexity of the incoming data points. For the benchmark, we assume perfect knowledge of which data points belong to which task and we assume full control over the data distribution. This trade-off has several advantages:

\begin{enumerate}
    \item As the tasks are arranged in increasing order of difficulty, it becomes much easier to quantify the change in the model's performance as the evaluation data distribution becomes different from the training data distribution.
    \item It enables us to extrapolate the capacity of the model with respect to the unseen tasks. If the model is unable to solve the $n^{th}$ task, it is unlikely to solve any of the subsequent tasks as they are harder than the current task. Thus, we can use the number of tasks solved (while keeping other factors like optimizer fixed) as an ordinal indicator of the model's capacity.
    \item As the data distribution becomes harder, the model is forced to use more and more of its capacity to learn the task.
    \item In general, given $n$ tasks, there are $n!$ ways of ordering the task and the model should be evaluated on all these combinations as the order of training tasks could affect the model's performance. Having the notion of the curriculum gives us a natural way to order the tasks. 
\end{enumerate}

To highlight the fact that curriculum-based training is not trivial, we show the performance of LSTM in the SSMNIST task in figure \ref{fig:seq}. We can see that training on different tasks makes the model highly susceptible to over-fitting to any given task and less likely to generalize across tasks.

Capacity saturation can happen because of two reasons in our proposed benchmark: 

\begin{enumerate}
\item The model is operating in a lifelong learning setting whereas the model learns a new task, it also needs to spend some capacity to retain knowledge about the previous tasks.
 
\item As the sequence length increases, the new tasks require more capacity to be learned.
\end{enumerate}

Given these factors, it is expected that as the model learns new tasks, its capacity would be strained, thus necessitating solutions that enable the model to increase its capacity on the fly.

\section{Model}

In this section, we first describe how the rehearsal setup is used in the GEM model and how the function preserving transformations can be used in the \textit{Net2Net} model. Next, we describe how we extend the \textit{Net2Net} model for RNNs. Then, we describe how the proposed model leverages both these mechanisms in a unified lifelong learning framework.

\subsection{Gradient Episodic Memory (GEM)}
\label{sec::model::gem}
In this section, we provide a brief overview of the Gradient Episodic Memory \citep{NIPS2017_7225} and how is it used for alleviating catastrophic forgetting while ensuring positive transfer on the backward tasks.

The basic idea is to store some input examples corresponding to each task (that the model has been trained on so far) in a memory buffer $B$. In practice, the buffer would have a fixed size, say $B_{size}$. If we know $T$, the number of tasks that the model would encounter, we could reserve $B_{size} / T$ number of slots for each task. Alternatively, we could start with the first task, use all the slots for storing the examples from the first task. Then, as we progress through tasks, we keep reducing the number of memory slots per task. While selecting the examples to store in the buffer, we just save the last few examples from each task. Specifically, we store only 1 minibatch of examples (10 examples) per task and find that even this small amount of data is sufficient.

As the model is training on the $l^{th}$ task, care is taken to ensure that the current gradient updates do not increase the loss on the examples already saved in the memory. This is achieved as follows: Given that the model is training on the $l^{th}$ task, we first compute the parameter gradient with respect to the data for the current task, which we denote as the \textit{current task gradient} or as $g_l$. Then a parameter gradient is computed corresponding to each of the previous tasks and is denoted as the \textit{previous task gradient}. If the current gradient $g_l$ increases the loss on any of the previous tasks, it is projected to the closest gradient $\tilde{g_l}$ (where closeness is measured in terms of $L_2$ norm) such that the condition is no more violated.  Whether the \textit{current task gradient} increases the loss on any of the previous tasks can be checked by computing the dot product between \textit{current task gradient} and the \textit{previous task gradient} (corresponding to the given previous task). The projected gradient update $\tilde{g_l}$ can be obtained by solving the following set of equations

\begin{align}
\text{minimize}_{\tilde{g_l}}  \quad \frac{1}{2} \quad& \|g_l - \tilde{g_l}\|_2^2\nonumber\\
\text{subject to} \quad& \langle \tilde{g_l}, g_k \rangle \geq 0 \text{ for all } k < l.\label{eq:gemprimal}
\end{align}

To solve \eqref{eq:gemprimal} efficiently, the authors use the primal of a Quadratic
Program (QP) with inequality constraints:
\begin{align}
    \text{minimize}_z \quad&\frac{1}{2} z^\top C z + p^\top z\nonumber\\
    \text{subject to} \quad&Az \geq b,\label{eq:primal}
\end{align}
where $C \in \mathbb{R}^{p \times p}$, $p\in \mathbb{R}^p$, $A \in
\mathbb{R}^{(t-1) \times p}$, and $b\in\mathbb{R}^{t-1}$.  The dual problem of~\eqref{eq:primal} 
is:
\begin{align}
    \text{minimize}_{u,v}   \quad&\frac{1}{2} u^\top C u - b^\top v\nonumber\\
    \text{subject to}       \quad&A^\top v - Cu = p,\nonumber\\
                                 &v \geq 0.\label{eq:dual}
\end{align}
If $(u^\star, v^\star)$ is a solution to \eqref{eq:dual}, then there is a
solution $z^\star$ to \eqref{eq:primal} satisfying $Cz^\star = Cu^\star$
\citep{dorn1960duality}.

The primal GEM QP \eqref{eq:gemprimal} can be rewritten as:
\begin{align*}
    \text{minimize}_z   \quad& \frac{1}{2} z^\top z -g^\top z + \frac{1}{2} g^\top g\\
    \text{subject to}   \quad& Gz \geq 0,
\end{align*}
where $G = -(g_1, \ldots, g_{t-1})$, and the constant term $g^\top g$ is discarded.
This new equation is a QP on $p$ variables (where $p$ is the number of parameters of the neural network). Since the network could have a lot of parameters, it is not feasible to solve this equation and the dual of the GEM QP is considered: 
\begin{align}
    \text{minimize}_{v}   \quad&\frac{1}{2} v^\top GG^\top v + g^\top G^\top v\nonumber\\
    \text{subject to} \quad&v \geq 0,\label{eq:project}
\end{align}
since $u = G^\top v + g$ and the term $g^\top g$ is constant.  This is a QP on
$t-1 \ll p$ variables (where $t$ is the number of observed tasks so far). Solution for the dual problem \eqref{eq:project}, $v^\star$, can be used to recover the projected gradient update as $\tilde{g} = G^\top v^\star  + g$.  The authors recommend adding a small constant $\gamma \geq 0$ to $v^\star$ as it helps to bias the gradient projection to updates that favoured beneficial backwards transfer.

We refer to this projection step as computing the \textit{GEM gradient} and the resulting update as the \textit{GEM update}. Since the projected gradient is only constrained to not increase the loss on the previous examples, a beneficial backward transfer is possible.

There are several downsides of using the GEM model. First, the projection of \textit{current task gradient} regularizes the model, thereby decreasing its effective capacity. This effect can be seen in figure \ref{fig::current-acc} where for all the three ``task distributions'', the green curve (Large LSTM model which does not use the GEM update) consistently outperforms the the red curve (LSTM model which uses the GEM update) both in terms of \textit{current task accuracy} and in terms of numbers of tasks completed. We counter this limitation by using the functional transformations to enable capacity expansion. Another downside is the cost - both in terms of computation and memory - of storing and rehearsing over the previous examples. We found that for all our experiments, storing just 10 examples per task is sufficient to get benefit from the GEM model. Hence the memory footprint of storing the training examples is very small and almost negligible as compared to the memory cost of persisting different copies of the model. The computational overhead of computing the GEM \textit{gradient} could be reduced to some extent by controlling the frequency at which the model rehearses on the previous examples and future work could look at a more systematic approach to eliminate or reduce this computational cost.

\subsection{Net2Net}

Training a lifelong learning system on a continual stream of data can be seen as training a model with an infinite amount of data. As the model experiences more and more data points, the size of its effective training dataset increases and eventually the network would have to expand its capacity to continue training. \textit{Net2Net} \citep{chen2015net2net} proposed a very simple technique, based on function preserving transformations, to achieve zero-shot knowledge transfer when expanding a small, trained network (referred to as the teacher network) into a large, untrained network (referred to as the student network). Given a teacher network represented by the function $y = f(x, \theta)$ (where $\theta$ refers to the network parameters), a new set of parameters $\phi$ are chosen such that $\forall x, f(x, \phi) = g(x, \theta)$. The paper considered two variants of this approach - \textit{Net2WiderNet} which increases the width of an existing network and \textit{Net2DeeperNet} which increases the depth of the existing network. The main benefit of using function-preserving transformations is that the student network immediately performs as well as the original network without having to go through a period of low performance. 

We use the \textit{Net2WiderNet} for expanding the capacity of the model. The \textit{Net2WiderNet} formulation is as follows: 

Assume that we start with a fully connected network where we want to widen layers $i$ and $i+1$. The weight matrix associated with layer $i$ is $\mW^{(i)} \in \sR^{m \times n}$ and that associated with layer $i+1$ is $\mW^{(i+1)} \in \sR^{n \times p}$. Layer $i$ may use any element-wise non-linearity. When we widen layer $i$, the weight matrix $\mW^{(i)}$ expands into $\mU^{(i)}$ to have $q$ output units where $q > n$. Similarly, when we widen layer $i+1$, the weight matrix $\mW^{(i+1)}$ expands into $\mU^{(i+1)}$ to have $q$ input units.

A random mapping function $g: \{1,2, \cdots, q\} \rightarrow \{1,2, \cdots, n\}$, is defined as: 
\[
g(j) = \left\{
\begin{array}{ll}
j & j \leq n\\
\mbox{ random sample from } \{1,2,\cdots n\} & j > n\\
\end{array}
\right.
\]

For expanding $\mW^{(i)}$, the columns of $\mU^{{(i)}}$ are randomly chosen from $\mW^{{(i)}}$ using $g$ as shown:
\[
\mU^{(i)}_{k,j} =  \mW^{(i)}_{k, g(j)}
\]

Notice that the first $n$ columns of $\mW^{(i)}$ are copied directly into $\mU^{(i)}$. 

The rows of $\mU^{{(i+1)}}$ are randomly chosen from $\mW^{{(i+1)}}$ using $g$ as shown:

\[
 \mU^{(i+1)}_{j,h} = \frac{1}{|\{x| g(x) = g(j)\}|} \mW^{(i+1)}_{g(j), h}   
\]

Similar to the previous case, the first $n$ rows of $\mW^{(i+1)}$ are copied directly into $\mU^{(i+1)}$. 

The replication factor, (given by $\frac{1}{|\{x| g(x) = g(j)\}|}$), is introduced to make sure that the output of the two models is exactly the same. This procedure can be easily extended to multiple layers. Similarly, the procedure can be used for expanding convolutional networks (where layers will have more convolution channels) as convolution is multiplication by a doubly block circulant matrix).

Once the training network has been expanded, the newly created larger network can continue training on the incoming data. In theory, there is no restriction on how many times the \textit{Net2Net} transformation is applied though we limit to using the transformation only once for most of our experiments. 

While \citet{chen2015net2net} mention lifelong learning as one of their motivations, they only focused on transfer learning from smaller network to a larger network for the single-task setup. Secondly, they considered the \textit{Net2Net} transformation in the context of feed-forward and convolutional models. Our work is the first attempt to use \NN~style function transformations for model expansion in the context of lifelong learning or even for sequential models.

\subsection{Extending Net2Net for RNNs}

In this section, we discuss the applicability of the \textit{Net2Net} formulation for the RNNs in the context of lifelong learning. 

The \textit{Net2WiderNet} transformation makes two recommendations about the training of the student network. The first is that the learning rate for the student network may be reduced by an order of 10. This argument seems useful in the original setup in which \textit{Net2Net} is proposed: training the student model over the same data on which the teacher model was trained. In the context of lifelong learning, the model does not see the same data again and the data distribution changes with the task. Hence the argument about lowering the learning rate does not apply. Our preliminary experiments showed that reducing the learning rate degrades the performance of the model. Hence we decided not to reduce the learning rate after the expansion.

The second and more important recommendation is that a small amount of random noise should be added to the student network to break the symmetry. In our initial experiments, we found that adding noise is a requirement and the model without noise performs extremely poor. This is in contrast to the feed-forward setting where the model works quite well even without using noise.

In the case of RNNs, when we apply the \textit{Net2Wider} transformation, the condition number of the hidden-to-hidden matrices increases drastically and it becomes ill-conditioned. Recall that the condition number is defined as the ratio of the largest singular value of the matrix to its smallest singular value. The ideal condition number would be 1 (as is the case of orthogonal matrices) and ill-conditioned networks are harder to train. Without adding noise, the condition number becomes infinity after expansion. This is due to the presence of correlated rows in the matrix. One way to get around this problem is to add a small amount of noise which helps to precondition the weight matrices and hence reduce their condition number. The issue with adding random noise is that it breaks down the equality condition and hence comes with a trade-off - A higher amount of random noise reduces the condition number more (make it better conditioned) but pushes the output of the newly instantiated student network away from the predictions of the old teacher network.

To that end, we propose a simple extension to the noise addition procedure which ensures that the output of the student and the teacher networks remain the same while taking care of the preconditioning aspect. Let us say that we had the weight matrix $\textbf{W}_{m \times n}$ which we expanded into $\textbf{U}_{m \times p}$ using the \textit{Net2WiderNet} transformation (where $p > n$). $\textbf{U}$ would have some columns of $\textbf{W}$ replicated. Let us say that the $i^{th}$ column was replicated $j$ times. Then, we would generate a noise matrix of small random values of size $m \times j$. The columns from this noise matrix would be added to columns that were replicated from $i^{th}$ column of the input matrix $W$. The noise matrix is generated such that for any row in the matrix, the sum of elements in that row of the noise matrix is 0. It can be shown mathematically that this transformation gives the exact same output as the case of no noise. We have to employ this procedure to make sure that the random noise we add sums up to 0. Since the given noise is random, it eliminates the correlation between rows and columns of the expanded weight matrix. Since the noise sums up to 0, it does ensure that the output of the student network is the same as that of the teacher network. 

How do we generate a matrix of random values where the sum of values along each row is 0? We describe a technique to generate a vector of random numbers such that the values sum up to 1 and then we can use the technique multiple times to sample multiple rows to form the matrix. Let us say we want to generate a vector of random values of length $k$ such that the values sum to 1. We first sample $k-1$ random points in the range (0, 1). Note that all these $k-1$ values will be smaller than $1$ and larger than $0$. We added the numbers $0$ and $1$ to this sequence and sort the sequence in the ascending order. This gives us a sorted sequence of $k+1$ points where each point lies in the range [0, 1] with the \textit{first value} being $0$ and the \textit{last value} being $1$. We take pairwise difference of values between the adjacent points \textit{i.e.} \textit{(second value - first value), (third value - second value)} and so on. Summing up this sequence of values would give us \textit{(last value - first value)} as all the other terms would cancel out. Since the \textit{first value} is $0$ and the \textit{last value} is $1$, the sum of the sequence of resulting $k$ points is $1$. From this sequence of numbers, we can subtract $1/k$ and the resulting sequence would exactly sum up to $0$. These steps are also described in Algorithm~\ref{alg::random}. Additionally, we scale the noise so that it is in the same range as the magnitude of the weights of the teacher network. Scaling the noise does not change the sum of the noise elements as both the positive and the negative elements get scaled by the same amount and still cancel each other. We use this strategy while using the expansion step.
\begin{algorithm}
\caption{Generating a random-valued vector of length $k$ where the values sum to 0}\label{alg::random}
\begin{algorithmic}[1]
\Procedure{Generator}{$k$}
\State Sample $k-1$ random points in the range ($0$, $1$).
\State Add values $0$ and $1$ to the sequence of sampled values.
\State Sort the sequence and create a new sequence by subtracting the pairwise values from the sorted sequence.
\State The resulting sequence of $k$ values will sum to $1$ (described in the text).
\State From each of the values, subtract $1/k$ to ensure that the resulting sequence of random values sums up to $0$.
\EndProcedure
\end{algorithmic}
\end{algorithm}

\subsection{Unified Model}

\label{sec::model::unified}

We now describe how we combine the catastrophic forgetting solution (GEM) and the capacity expansion solution (functional transformations) to come up with a more suitable model for lifelong learning. Given a ``task distribution'', we randomly initialize a model, reset the episodic memory to be empty and start training the model on the first task (simplest task). Once a task is learned, the model starts training on the subsequent, more difficult tasks. When we are training the model on the $l^{th}$ task, the episodic memory already has some examples corresponding to the first $l-1$ tasks. The \textit{current task gradient} is projected with respect to the \textit{previous task gradient}s to ensure that it does not increase the loss associated with any of the examples in the episodic memory. The projected \textit{GEM Gradient} is used to update the weights of the model (\textit{GEM Update}). The model is trained on the current task for a fixed number of iterations. The last $m$ training examples from the current task are stored in the episodic memory for use in the subsequent tasks. In general, the $m$ examples can be selected with some more sophisticated strategy though \cite{NIPS2017_7225} reports, and we validate, that using just the last $m$ samples works well in practice.

If the model complete learning the current task (i.e. achieves a threshold amount of accuracy after training), the model can start training on the next task. If the model fails to learn the current task, and has not been expanded so far, the model is expanded to a larger model and is allowed to train further on the current task. Once the expanded model is trained, it is re-evaluated to check if it has learned the task. If it has, the model progresses to the next task, otherwise, the training procedure is terminated. Irrespective of how much is the \textit{current task accuracy}, the model is evaluated on all the tasks - to measure its \textit{previous task accuracy} and \textit{future task accuracy}.

\subsection{Analysis of the computational and memory cost of the proposed model}
\label{sec::model::perf}

As noted in section~\ref{sec::introduction}, an important desideratum in lifelong learning models is that the computational and the memory costs of the model should ideally grow sublinearly as the model is trained on new tasks. In the context of our proposed model, the computational and memory costs can change in the following ways:

\begin{enumerate}
    \item The \textit{Net2Net} component expands the model. In this case, the expanded model would take more resources (both in terms of parameters and time) than the earlier model. We note that the expansion step does not happen for every new task and is performed only when the model's capacity saturates. This is in contrast to approaches like \cite{rusu2016progressive} where a new copy of the network is added every time a new task is introduced thus increasing both the parameters and the compute time linearly with the number of tasks. In our case, the frequency of expansion is sublinear in the number of tasks.

    \item The GEM model stores some examples (in a buffer) from the previous tasks and performs gradient computation with respect to those examples, along with the gradient computation for the current examples. As noted in section~\ref{sec::model::gem}, we could keep the buffer size to be fixed and replace some examples from the previous tasks as new examples are observed while making sure that all the tasks are represented through examples in the buffer. In practice, we found that storing a few examples per task is sufficient to get benefit from the GEM model (as also observed by~\cite{NIPS2017_7225}), making the memory footprint negligible. As noted earlier, future work could look at some systematic ways of selecting the examples from the buffer thus reducing the computational overhead.
    
\end{enumerate}

One beneficial side effect of using \textit{Net2Net} expansion is the zero-shot knowledge transfer that further amortizes the cost of training a newly initialized larger model - either from a smaller, pre-trained model or from dataset corresponding to the tasks encountered so far.

\section{Experiments}
\label{sec::experiment}

\subsection{Models}
\label{subsec:Model}

For each ``task distribution'', we consider a standard recurrent (LSTM) model operating in the lifelong learning setting. We consider the different aspects of training a lifelong learning system and describe how the model variants can account these aspects.
We start with an LSTM model with hidden state size of 128 and refer to this model as the \textit{small-Lstm} model. This model has sufficient capacity to learn the first few tasks.  We start training the  (\textit{small-Lstm}) model as described in section \ref{sec::task::benchmark}. To avoid catastrophic forgetting, we could additionally use the \textit{GEM update} when training the model. The resulting model is referred to as the \textit{small-Lstm-Gem} model. After learning some tasks, the model would have used up all its capacity (since it is retaining the knowledge of the previous tasks as well). In this case, we could expand the model's capacity using the \textit{Net2Net} transformation and the model with this capability is referred to as the  \textit{small-Lstm-Gem-Net2Net} model. This is the model we propose. Alternatively, we could have started the training with a larger model (\textit{large-Lstm} model) and could have used the GEM update (\textit{large-Lstm-Gem} model) to counter forgetting. The strategy of always starting training with a large network would not work in practice because in the lifelong learning setting we do not know what network would be sufficiently large to learn all the tasks beforehand. If we start with a very large model, we would need a lot more computational resources to train the model and the model would be very prone to over-fitting. Our proposed model  (\textit{small-Lstm-Gem-Net2Net}) gets around this problem by increasing the capacity on the fly as and when needed. For the \textit{large-Lstm} model family, we set the size of the hidden layer to be 256. Our empirical analysis shows that it is possible to expand the models to a size much larger than their current size without interfering with the \textit{GEM update}.

For the performance on the current task, \textit{large-Lstm} model can be treated as the gold standard since this model has the largest capacity among all the models considered. Unlike the models which use the \textit{GEM Update}, this model does not have to ``use'' some of its capacity for retaining the knowledge of the previous tasks. For the performance on the previous tasks (catastrophic forgetting), we consider the \textit{large-Lstm-Gem} model as the gold standard as this model has the largest capacity among all the models and is specifically designed to counter catastrophic forgetting. While we do not have a gold standard for the \textit{Future Task Accuracy}, both \textit{large-Lstm-Gem} and \textit{large-Lstm} are reasonable models to compare with. Overall, we have three different gold standards for three setups (and metrics) and we compare our proposed model to these different gold standards (each specialized for a specific use-case).

\subsection{Hyper Parameters}
\label{subsec::hyper-param}

All the models are implemented using PyTorch 0.4.1 \citep{pytorch}. Adam optimizer \citep{adam} is used with a learning rate of 0.001. We used one layer LSTM models with hidden dimensions of size 128 and 256. \textit{Net2Net} is used to expand LSTM models of size 128 to 256. For the GEM model, we keep one minibatch (10 examples) of data per task for obtaining the projected gradients. We follow the guidelines and hyperparameter configurations as specified in the respective papers for both \textit{GEM} and \textit{Net2Net} models.

\begin{figure*}[h!]
  \begin{subfigure}{\columnwidth}
  {\includegraphics[width=0.32\columnwidth]{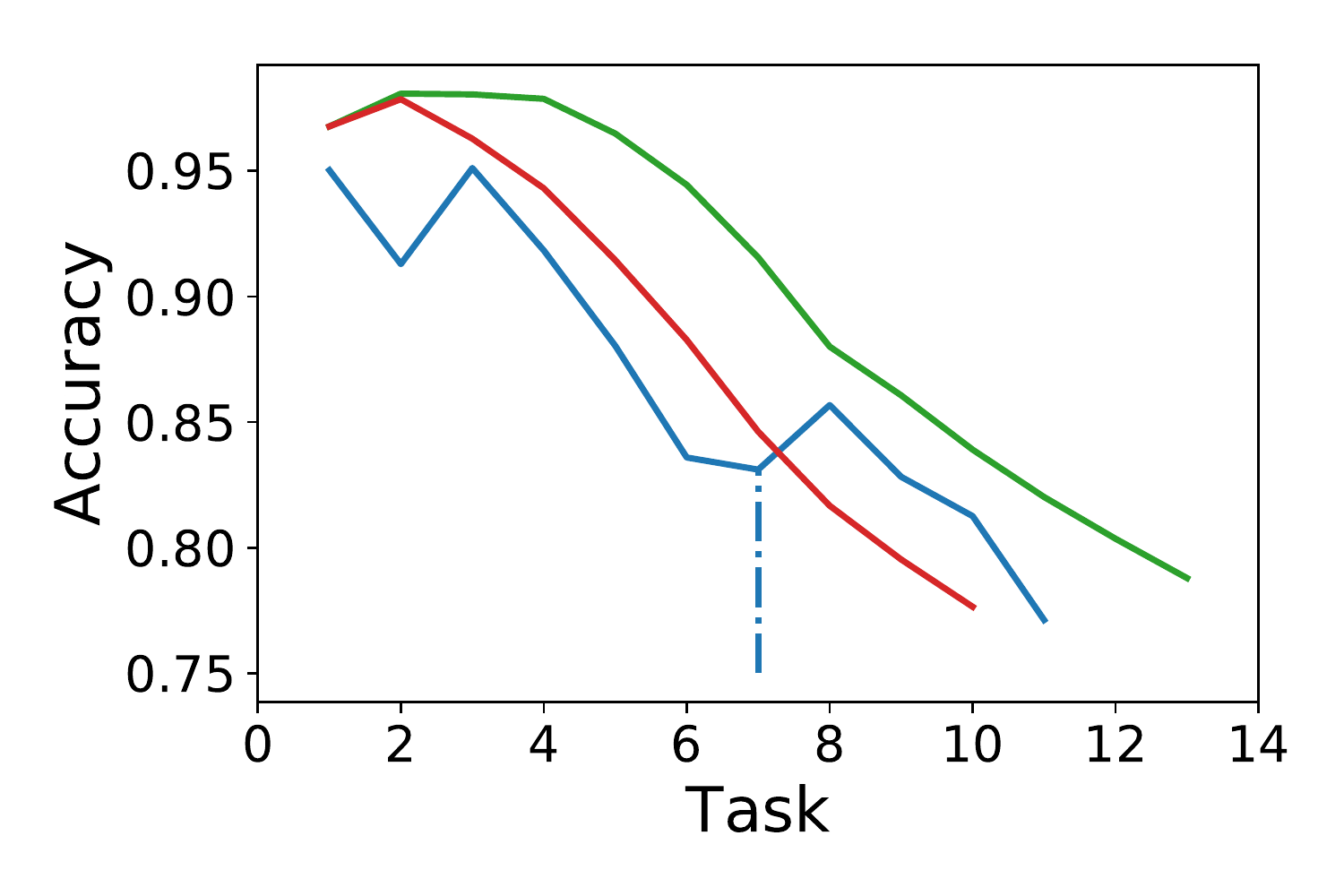}}
  {\includegraphics[width=0.32\columnwidth]{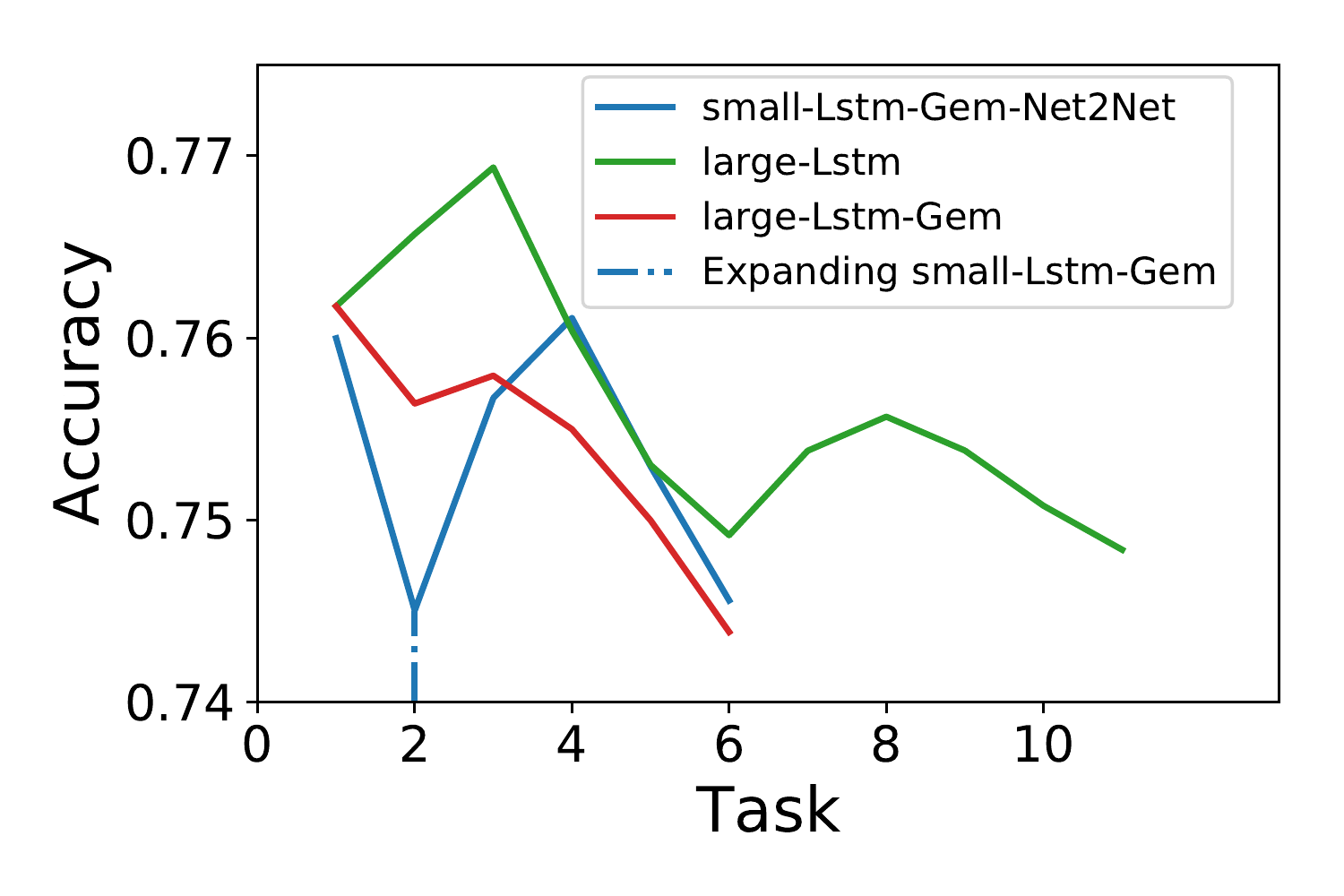}}
  {\includegraphics[width=0.32\columnwidth]{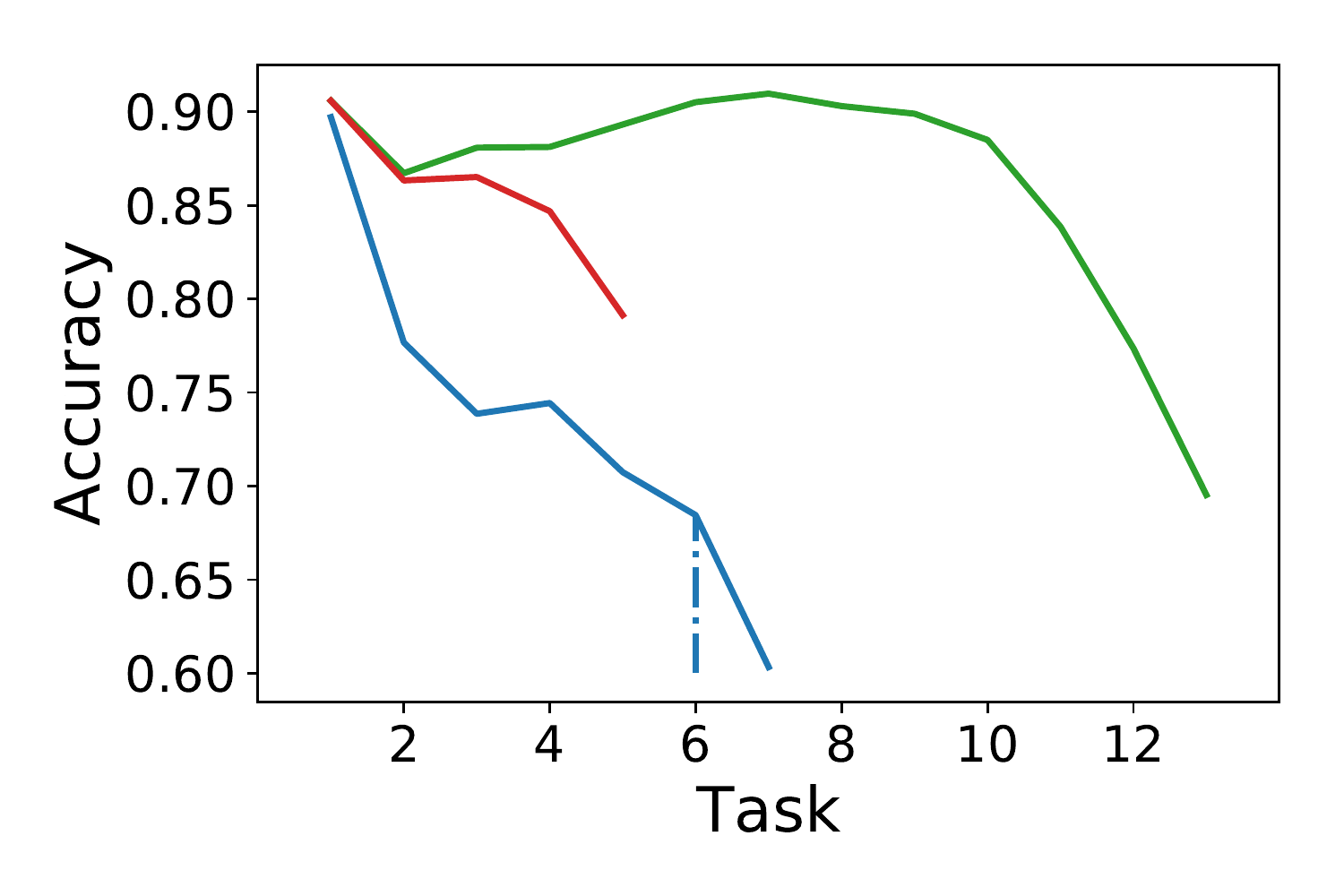}}
  \end{subfigure} 
  \caption{\textit{Current Task Accuracy} for the different models on the three ``task distributions'' (Copy, Associative Recall, and SSMNIST respectively). On the x-axis, we plot the index of the task on which the model is training currently and on the y-axis, we plot the accuracy of the model on that task. Higher curves have higher \textit{current task accuracy} and curves extending more have completed more tasks. For all the three ``task distributions'', our proposed  \textit{small-Lstm-Gem-Net2Net} model clears either more levels or same number of levels as the \textit{large-Lstm-Gem} model. Before the blue dotted line, the proposed model is of much smaller capacity (hidden size of 128) as compare to other two models which have a larger hidden size (256). Hence the larger models have better accuracy initially. Capacity expansion technique allows our proposed model to clear more tasks than it would have cleared otherwise.}
  \label{fig::current-acc}
\end{figure*}

\subsection{Results}
\label{subsec::results}

Figure \ref{fig::current-acc} shows the trend of the \textit{current task accuracy} for the different models on the three ``task distributions''. In these plots, a higher curve corresponds to the model that has higher accuracy on the current task and models which learn more tasks are spread out more along the x-axis. We compare the performance of the proposed model \textit{small-Lstm-Gem-Net2Net} with the gold standard \textit{large-Lstm} model. We additionally compare with \textit{large-Lstm-Gem} model as both this model and the proposed model are constrained to use some of their capacity on the previous tasks. Hence it provides a more realistic estimate of the strength of the proposed model. It also allows us to study the effect of the \textit{GEM Update} on the model's effective capacity (in terms of the number of tasks cleared). The blue dotted line corresponds to the expansion step when the model is not able to learn the current task and had to expand. This shows that using the capacity expansion technique from \textit{Net2Net} enables learning on newer tasks. We highlight that before expansion, the proposed model \textit{small-Lstm-Gem-Net2Net} had a much smaller capacity (128 hidden dims) as compared to the other two models which started with a much larger capacity (256 hidden dims). This explains why the larger models have much better performance in the initial stages. Post expansion, the proposed model overtakes the GEM based model in all the cases (in terms of the number of tasks solved). We can observe that in all the cases, \textit{large-Lstm} model outperforms the \textit{large-Lstm-Gem} model which suggests that using the \textit{Gem Update} comes at the cost of reducing the capacity for the current task. Using capacity expansion techniques with \textit{GEM} enables the model to account for this loss of capacity.

Figure \ref{fig::prev-acc} shows the trend of the \textit{previous task accuracy} for the different models. A higher bar corresponds to better accuracy on the previous tasks (more resilience to catastrophic forgetting). We compare the performance of the proposed model \textit{small-Lstm-Gem-Net2Net} with the gold standard model \textit{large-Lstm-Gem}. We additionally compare with the \textit{large-Lstm} model to demonstrate that \textit{GEM Update} is essential to have a good performance on the previous tasks. The most important observation is the relative performance of the proposed \textit{small-Lstm-Gem-Net2Net} model and the \textit{large-Lstm-Gem} model. The \textit{small-Lstm-Gem-Net2Net} model started as a smaller model, consistently learned more tasks than \textit{large-Lstm-Gem} model and is still almost as good as \textit{large-Lstm-Gem} model in terms of \textit{Previous Task Accuracy}. This shows that the proposed model is very robust to catastrophic forgetting while being very good at learning the current task. We also observe that for all the three ``task distributions'', the models using the \textit{GEM update} are more resilient to catastrophic forgetting as compared to the models without the \textit{GEM Update}.

\begin{figure*}[h]
  \begin{subfigure}{\columnwidth}
  {\includegraphics[width=0.32\columnwidth]{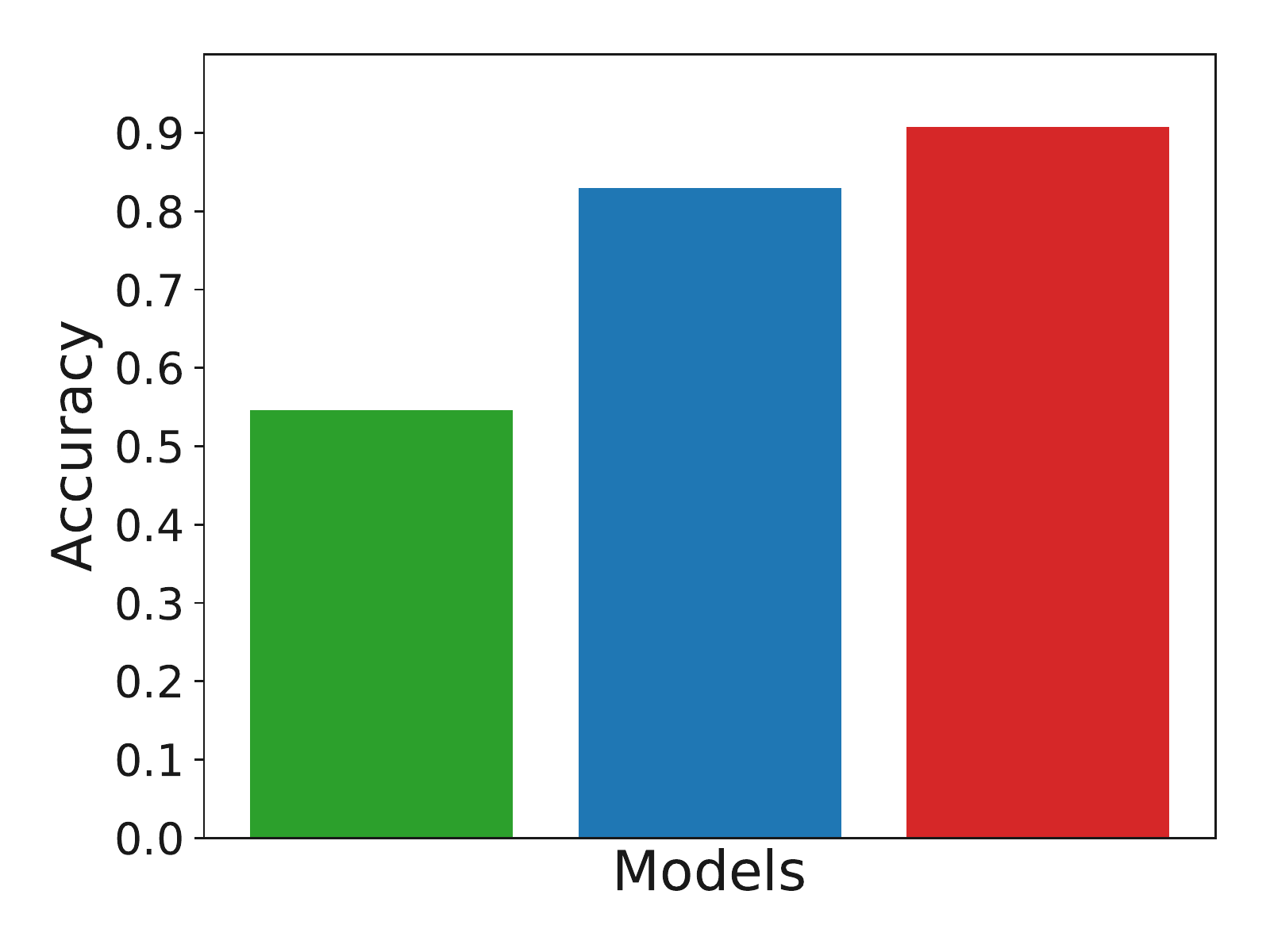}}
  {\includegraphics[width=0.32\columnwidth]{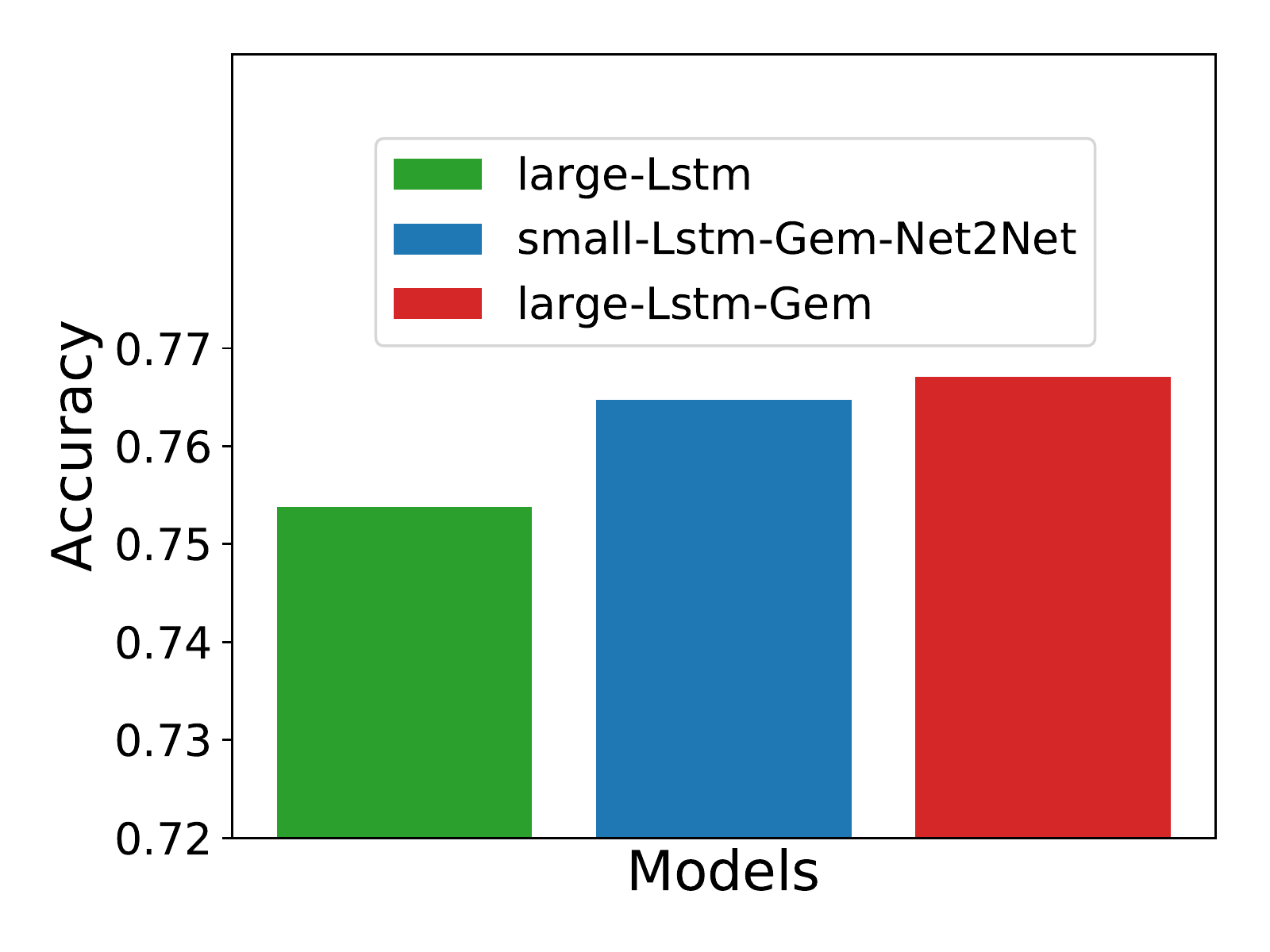}}
  {\includegraphics[width=0.32\columnwidth]{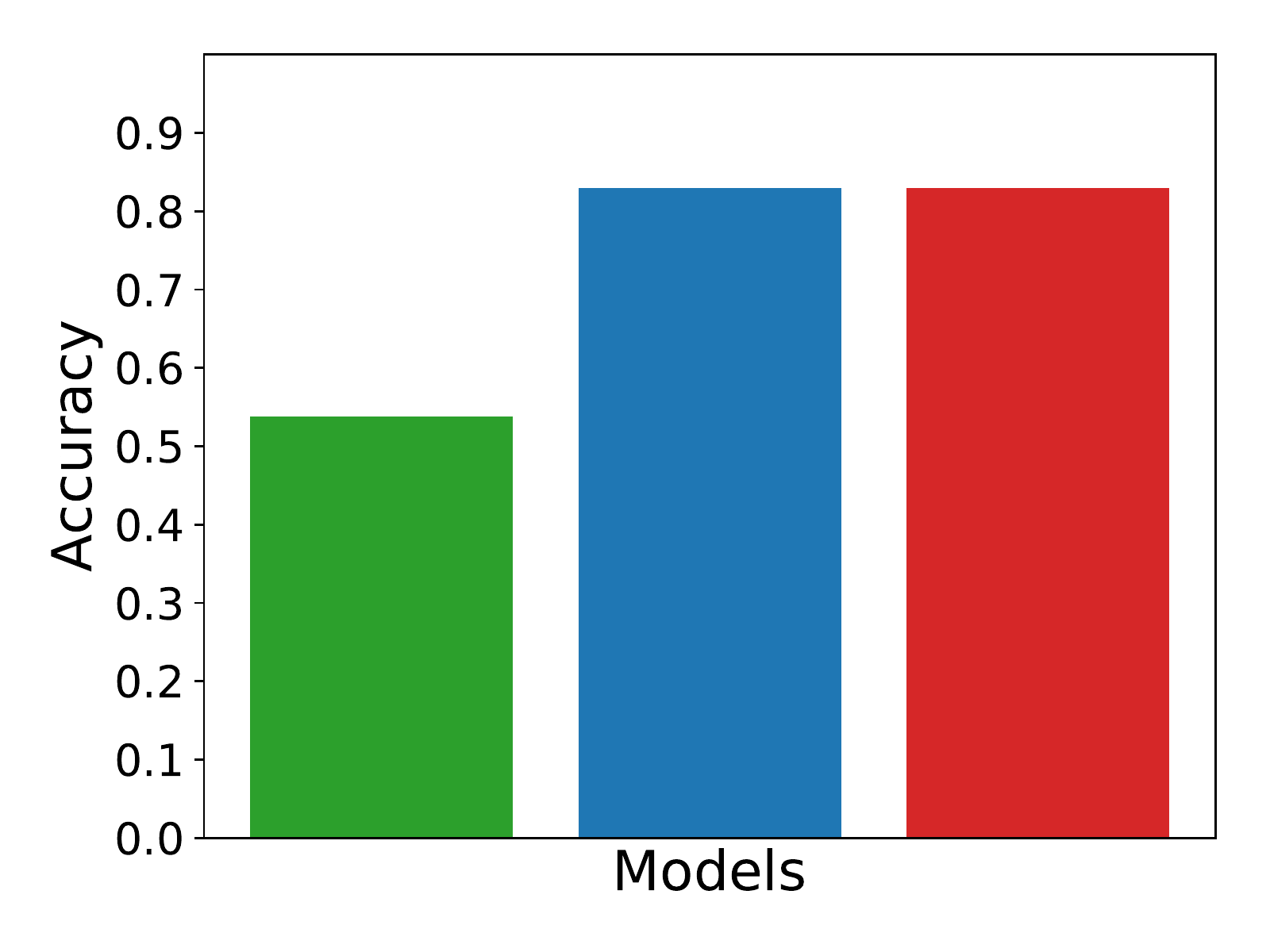}}

  \end{subfigure}\par\medskip
  \caption{\textit{Previous Task Accuracy} for the different models on the three ``task distributions'' (Copy, Associative Recall, and SSMNIST respectively). Different bars represent different models and on the y-axis, we plot the average {previous task accuracy} (averaged for all the tasks that the model learned). Higher bars have better accuracy on the previously seen tasks and are more robust to catastrophic forgetting. For all the three ``task distributions'', the proposed models are very close in performance to the \textit{large-Lstm-Gem} models and much better than the \textit{large-Lstm} models.}
  \label{fig::prev-acc}
\end{figure*}

\begin{figure*}[h]
  \begin{subfigure}{\columnwidth}
  {\includegraphics[width=0.32\columnwidth]{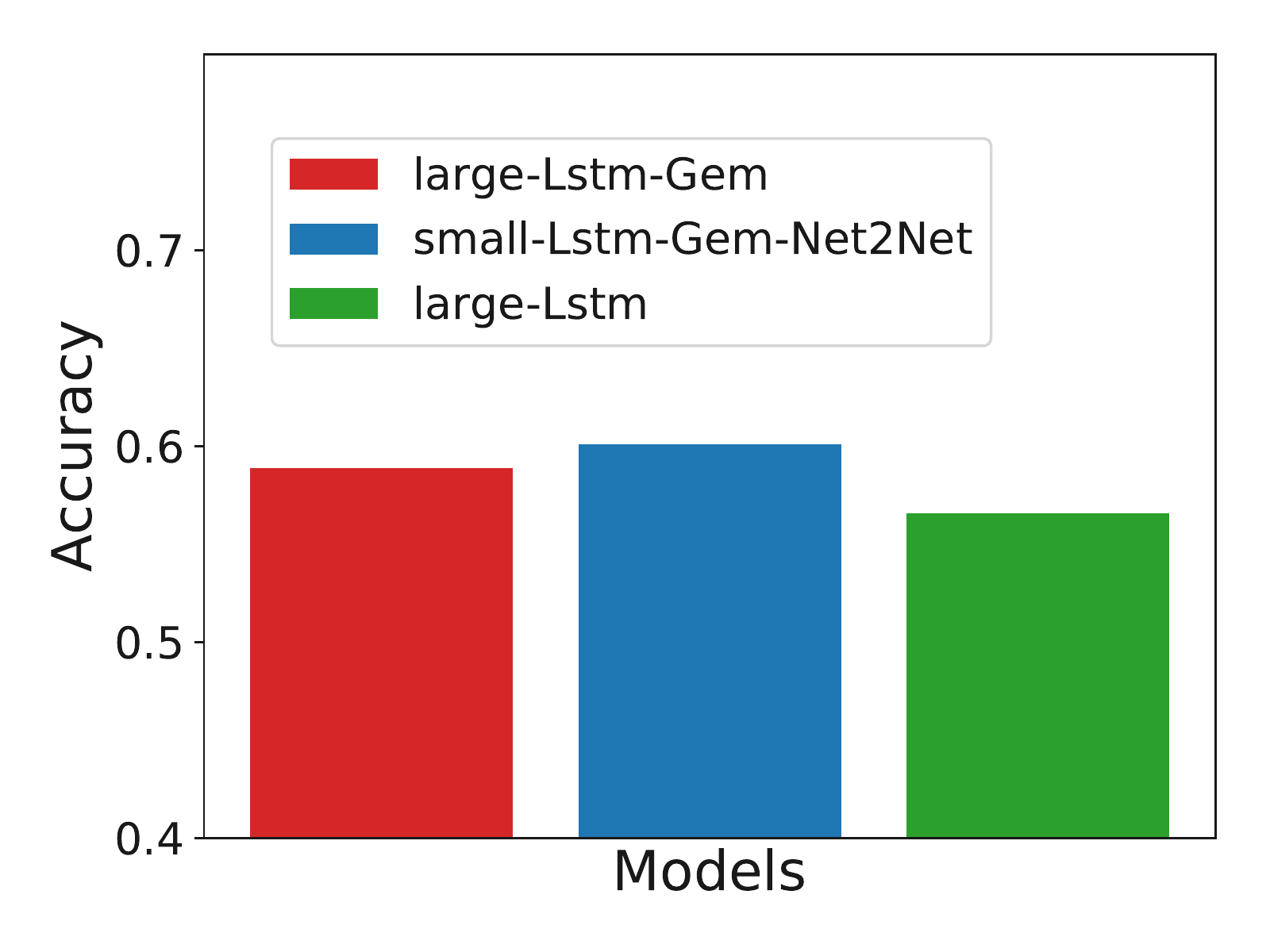}}
  {\includegraphics[width=0.32\columnwidth]{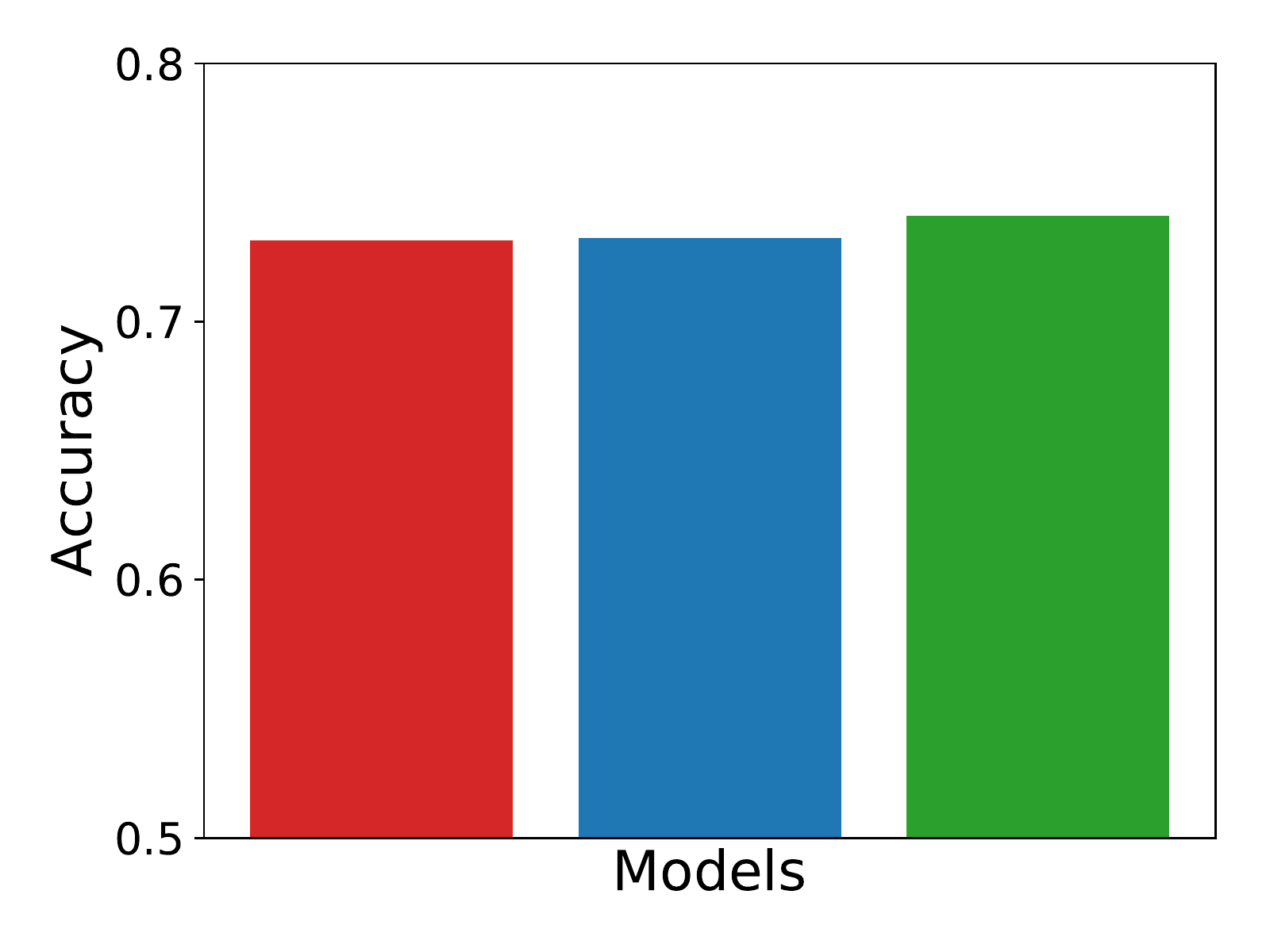}}
  {\includegraphics[width=0.32\columnwidth]{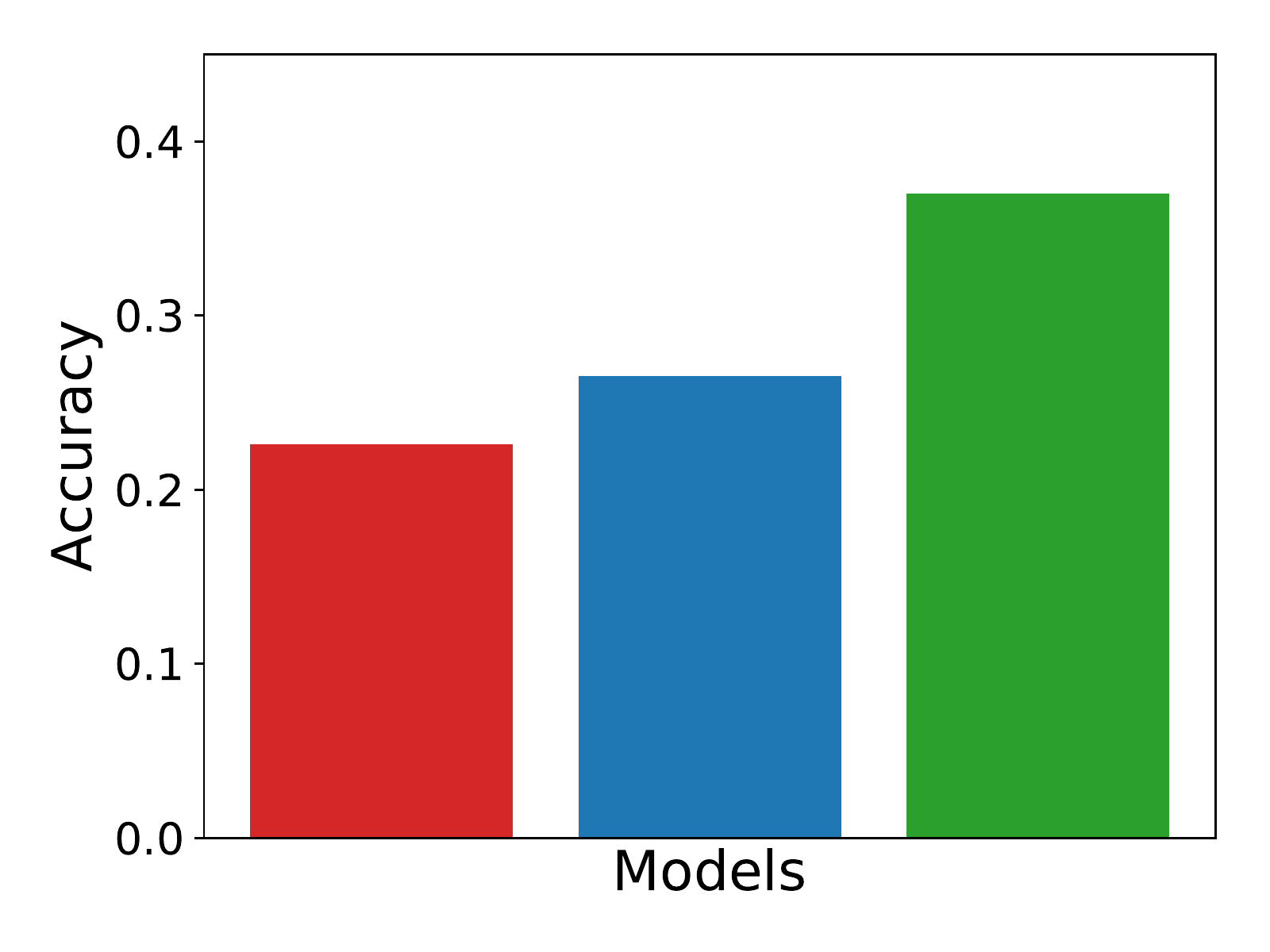}}
  \end{subfigure}\par\medskip
  \caption{\textit{Future Task Accuracy} for the different models on the three ``task distributions'' (Copy, Associative Recall, and SSMNIST respectively). Different bars represent different models and on the y-axis, we plot the average {future task accuracy} (averaged for all the tasks that the model learned). Higher bars have better accuracy on the previously unseen tasks and are more beneficial for achieving knowledge transfer to future tasks. Even though the proposed model does not have any component for specifically generalizing to the future tasks, we expect the proposed model to generalize at least as well as the \textit{large-Lstm-Gem} model and comparable to \textit{large-Lstm}. Interestingly, our model outperforms the \textit{large-Lstm} model for Copy task and is always better than (or as good as) the \textit{large-Lstm-Gem} model. }
  \label{fig::next-acc}
\end{figure*}

Figure \ref{fig::next-acc} shows the trend of the \textit{future task accuracy} for different models. A higher bar corresponds to better accuracy on the future (unseen) tasks. Since we do not have any gold standard for this setup, we consider both \textit{large-Lstm-Gem} and \textit{large-Lstm} models as they both are reasonable models to compare with. The general trend is that our proposed model is quite close to the reference models for 2 out of 3 tasks. Note that both the larger models started training with a much larger capacity and further, the \textit{large-Lstm} model is not constrained by the \textit{GEM Update} and hence the maximum amount of effective capacity. This could be one reason why the model can outperform our proposed model for one of the tasks. 

\begin{figure*}[htbp]
  \begin{subfigure}{\columnwidth}
  {\includegraphics[width=0.32\columnwidth]{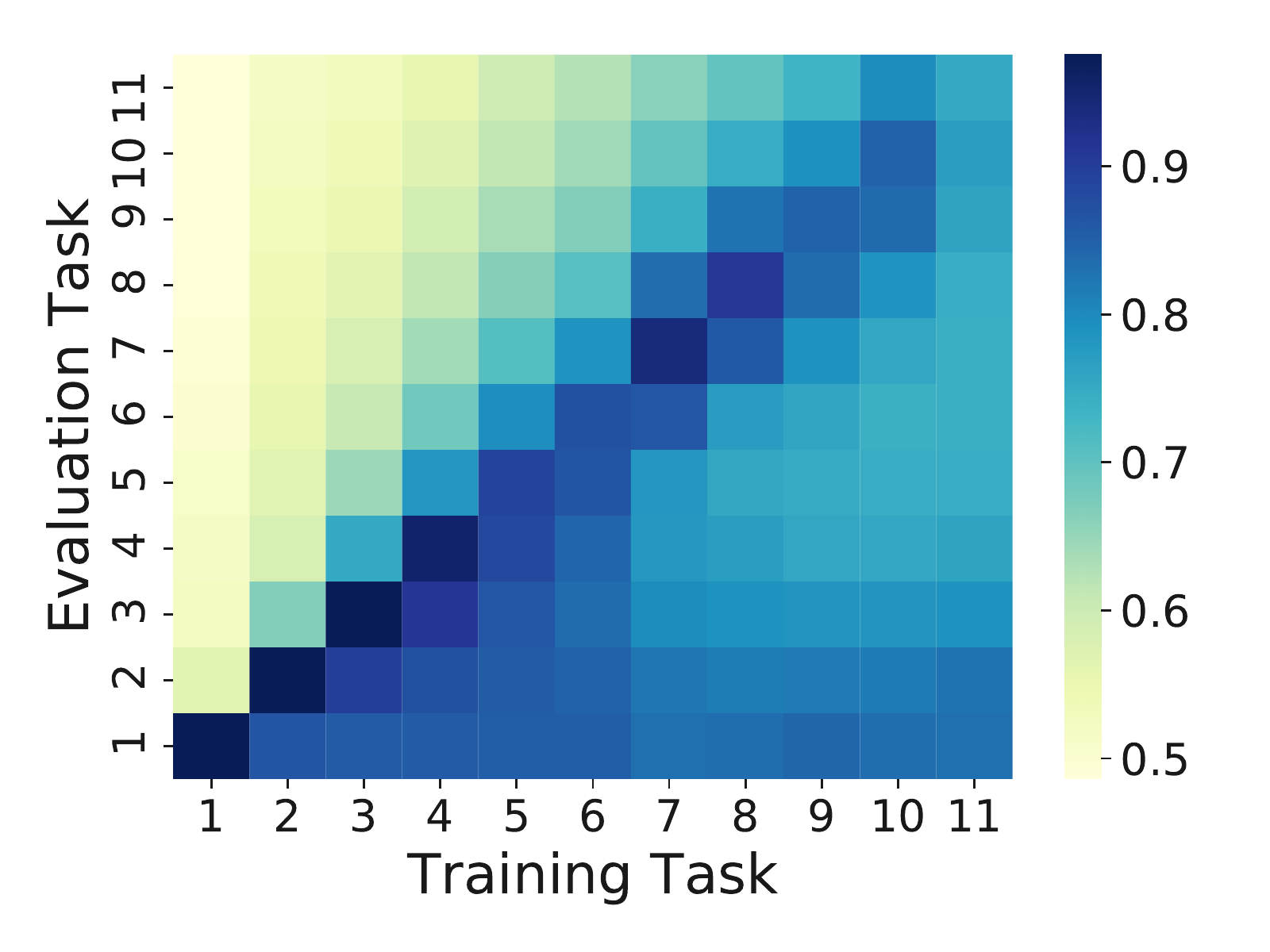}}
  {\includegraphics[width=0.32\columnwidth]{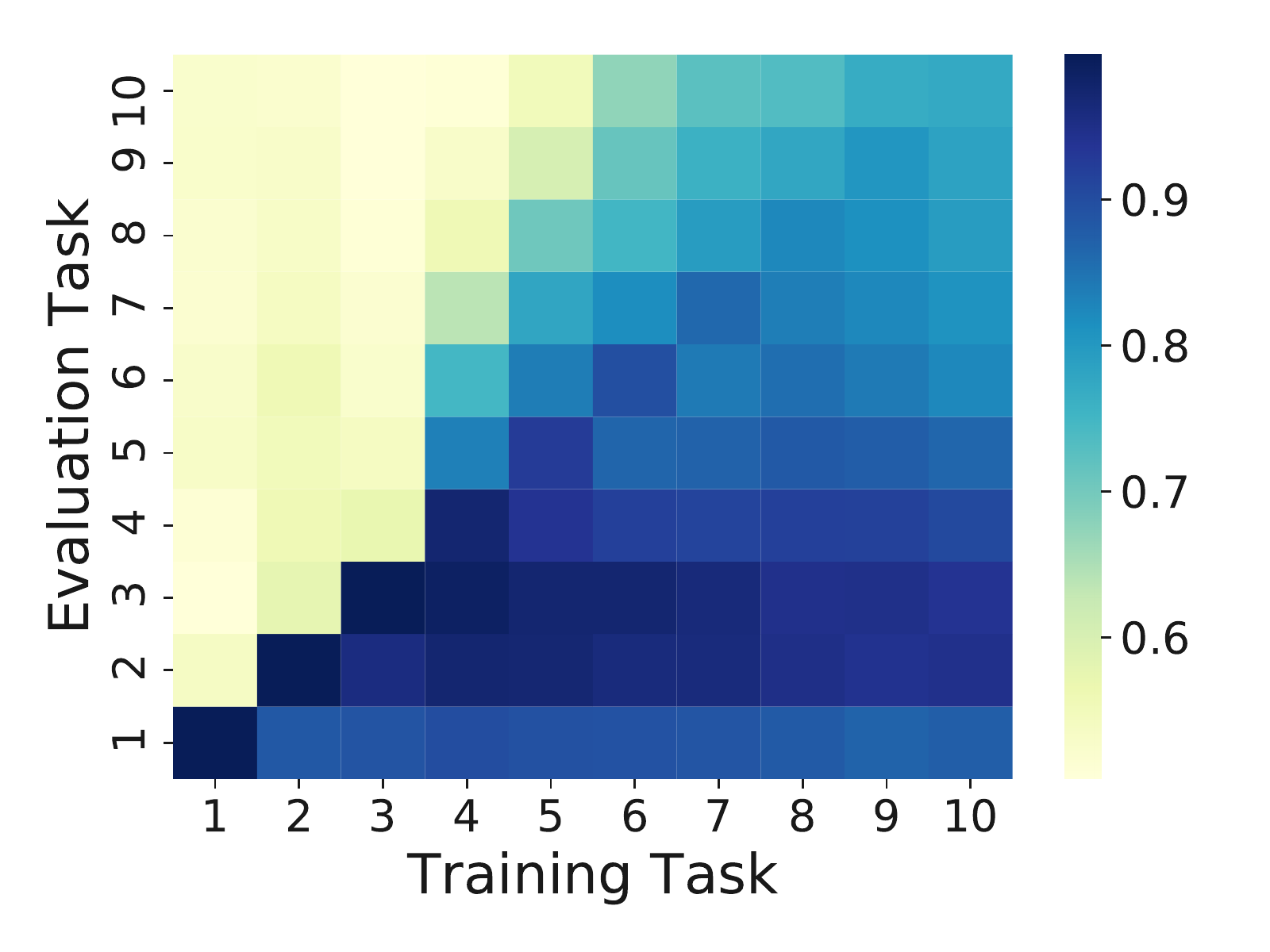}}
  {\includegraphics[width=0.32\columnwidth]{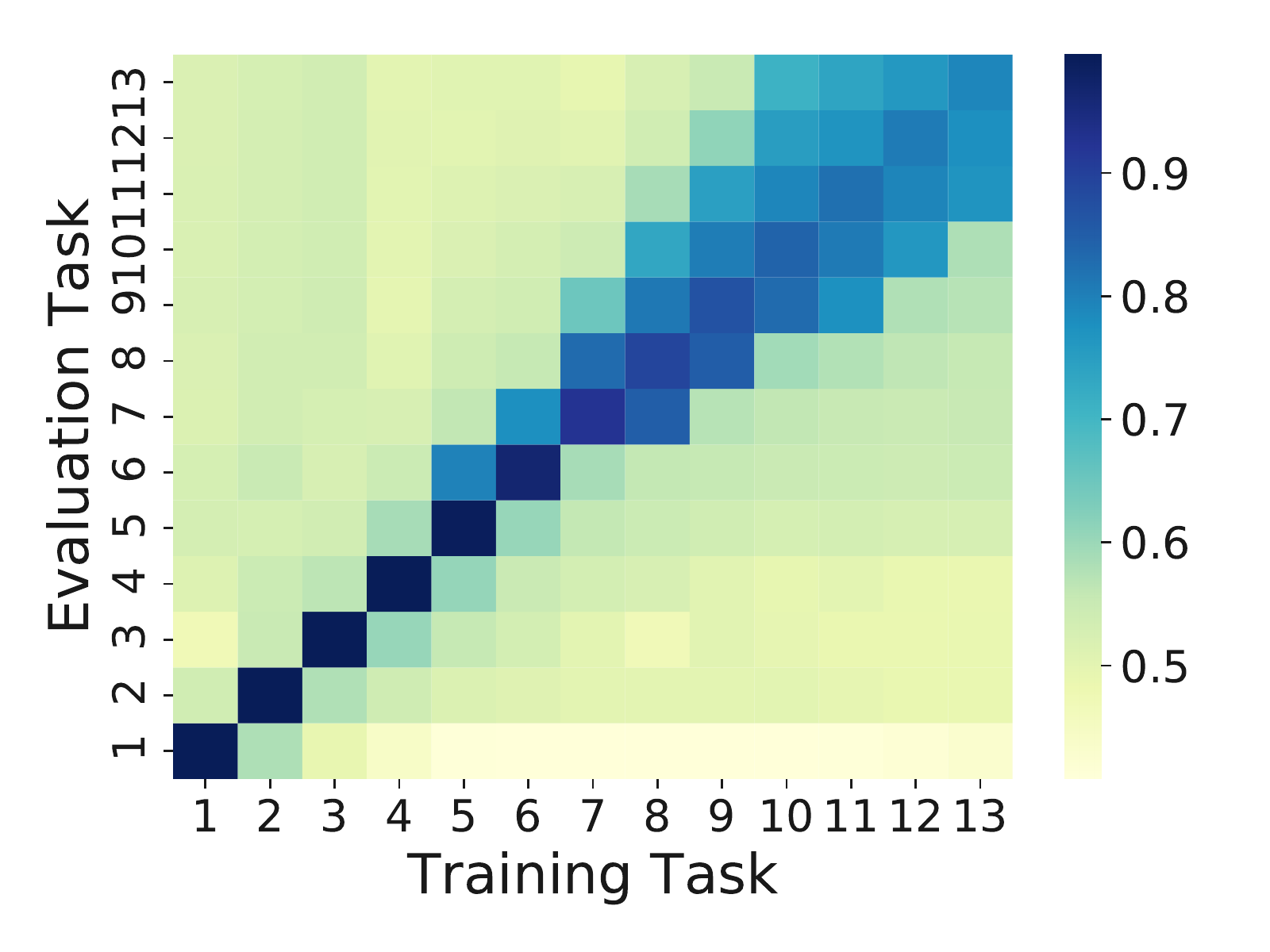}}
  \caption{Copy Task}
  \end{subfigure}\par\medskip
  \begin{subfigure}{\columnwidth}
  {\includegraphics[width=0.32\columnwidth]{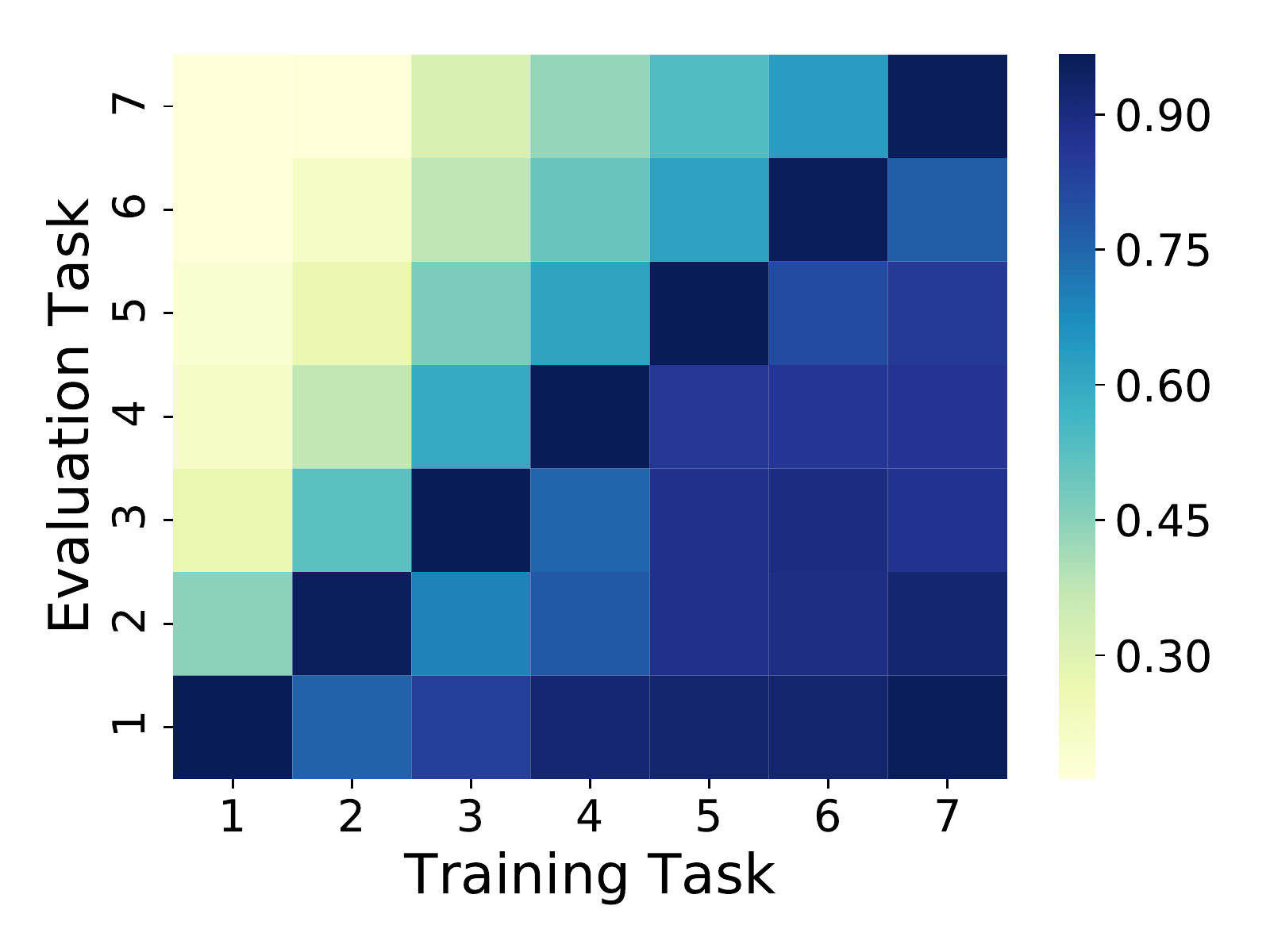}}
  {\includegraphics[width=0.32\columnwidth]{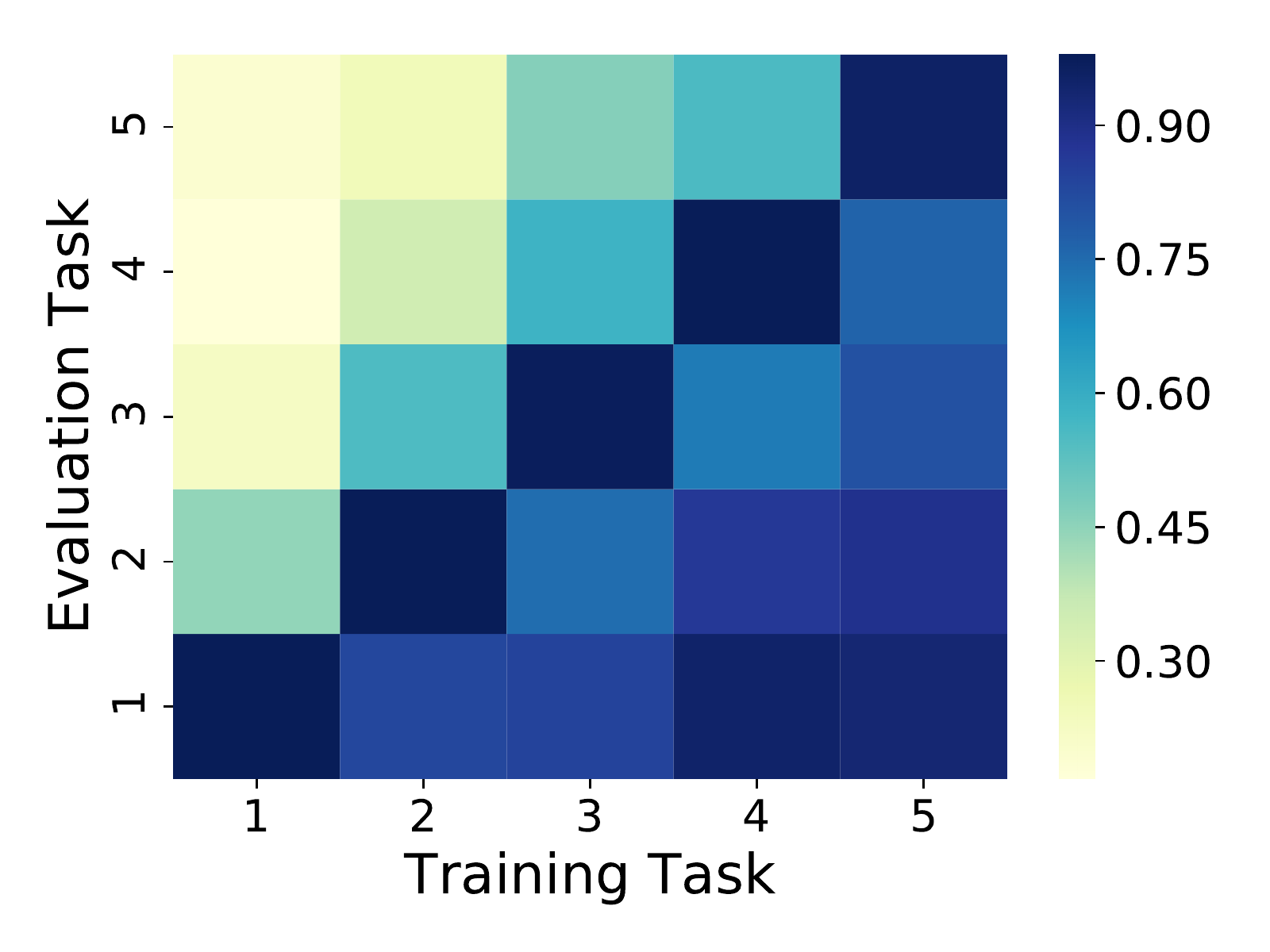}}
  {\includegraphics[width=0.32\columnwidth]{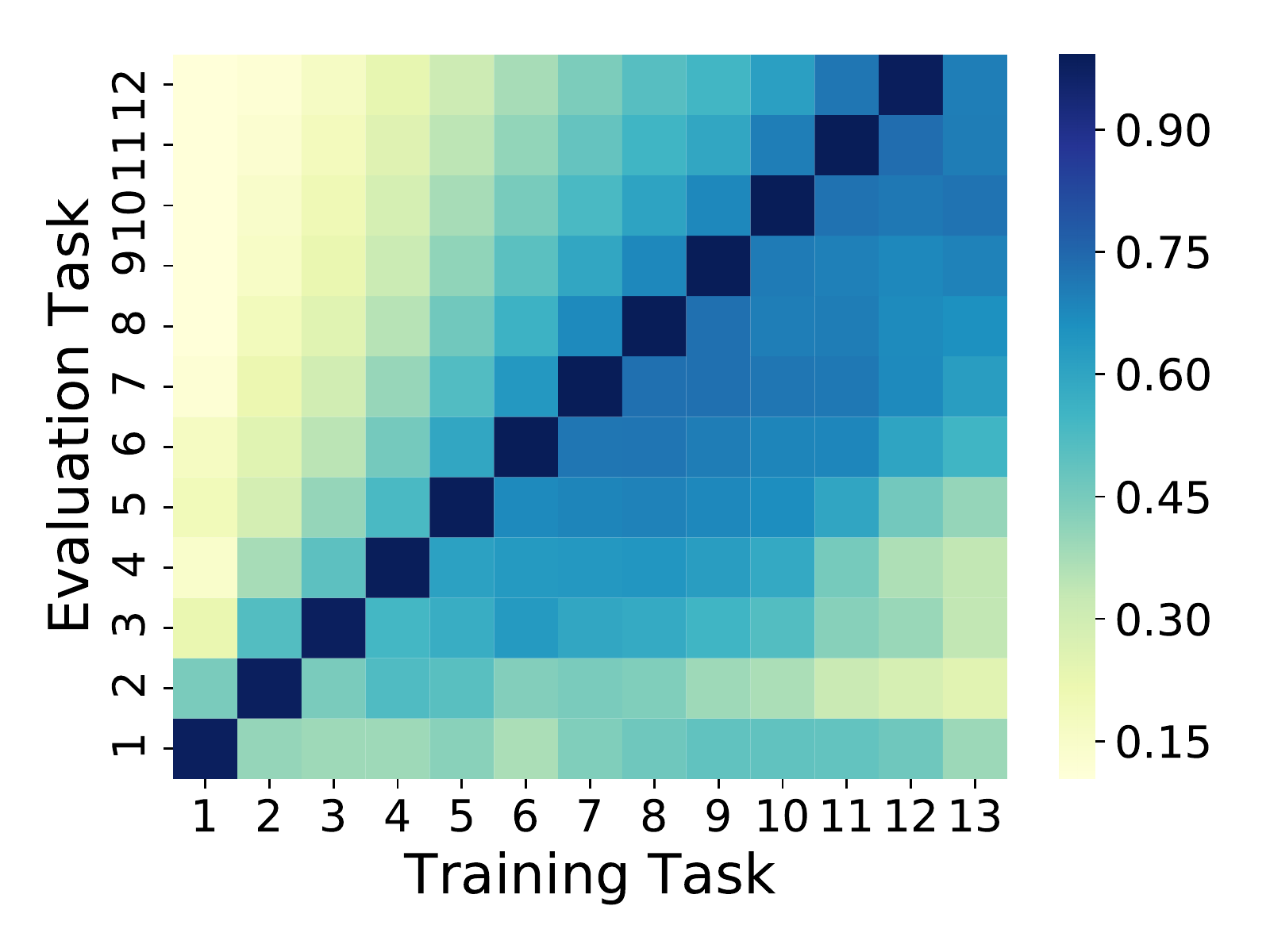}}
  \caption{SSMNIST Task}
  \end{subfigure}\par\medskip
  \caption{Accuracy of the different models (\textit{small-Lstm-Gem-Net2Net}, \textit{large-Lstm-Gem} and \textit{large-Lstm} respectively) as they are trained and evaluated on different tasks for the Copy and the SSMNIST ``task distributions''. On the x-axis, we show the task on which the model is trained and on the y-axis, we show the accuracy corresponding to the different tasks on which the model is evaluated. We observe that for the \textit{large-Lstm} model, the high accuracy values are concentrated along the diagonal which indicates that the model does not perform well on the previous task. In case of both \textit{small-Lstm-Gem-Net2Net} and \textit{large-Lstm-Gem} model, the high values are in the lower diagonal region indicating that the two models are quite resilient to catastrophic forgetting.}
  \label{fig::heat::copy}
\end{figure*}

We also consider the heatmap plots where we plot the accuracy of different models (for different ``task distributions'') as they are trained and evaluated on different tasks. As pointed out in section \ref{sec::task::benchmark}, the aggregated metrics (\textit{current task accuracy}, \textit{previous task accuracy}, etc) are not sufficient to compare the performance of different models and fine-grained analysis is useful for having a holistic view. We observe that for the \textit{large-Lstm} model, the large values are concentrated along the diagonal while for the \textit{small-Lstm-Gem-Net2Net} and the \textit{large-Lstm-Gem} models, the high values are concentrated in the lower diagonal region indicating that the two models are quite resilient to catastrophic forgetting. Additionally, note that while the \textit{large-Lstm-Gem} model appears to be more resilient to catastrophic forgetting, the \textit{small-Lstm-Gem-Net2Net} model consistently clears more tasks. Note that even though we are evaluating the models for all the tasks in the benchmark, we are restricting the heatmap to only show evaluation results for the highest task index that the model could solve. This results in square-shaped heatmaps which are easier to analyze. 

It is important to note that we are using a single proposed model (\textit{small-Lstm-Gem-Net2Net}) and comparing it with gold-standard models in 3 different contexts - performance on the current task, performance on the backward tasks and performance on the future tasks. Our model can provide strong performance on all the three tasks by countering catastrophic forgetting and by using capacity expansion.

\section{Conclusion}

In this work, we study the problem of capacity saturation and catastrophic forgetting in lifelong learning in the context of sequential supervised learning. We propose to unify  Gradient Episodic Memory (a catastrophic forgetting alleviation approach) and Net2Net (a capacity expansion approach) to develop a model that is more suitable for lifelong learning. We also propose a curriculum-based evaluation benchmark where the models are trained on a task with increasing levels of difficulty. This enables us to sidestep some of the challenges that arise when studying lifelong learning. We conduct experiments on the proposed benchmark tasks and show that the proposed model is better suited for the lifelong learning setting as compared to the two individual models. As future work, we would want to address the computational overhead associated with the \textit{GEM Update} step.

\bibliographystyle{apacite}
\bibliography{main}

\section{Appendix}

\subsection{Results}

\subsubsection*{Current Task Accuracy}

\begin{table}[H]
\begin{tabular}{|c|c|c|c|}
\hline
\textbf{Task Id} & \textbf{small-Lstm-Gem-Net2Net} & \textbf{large-Lstm} & \textbf{large-Lstm-Gem} \\ \hline
1                & 95.03                           & 96.77               & 96.77                   \\ \hline
2                & 91.29                           & 98.07               & 97.85                   \\ \hline
3                & 95.11                           & 98.04               & 96.283                  \\ \hline
4                & 91.83                           & 97.86               & 94.308                  \\ \hline
5                & 88.03                           & 96.48               & 91.446                  \\ \hline
6                & 83.58                           & 94.44               & 88.27                   \\ \hline
7                & 83.107 *                        & 91.55               & 84.61                   \\ \hline
8                & 85.67                           & 88.00               & 81.66                   \\ \hline
9                & 82.82                           & 86.06               & 79.52                   \\ \hline
10               & 81.25                           & 83.90               & 77.64                   \\ \hline
11               & 77.12                           & 82.01               &                         \\ \hline
12               &                                 & 80.35               &                         \\ \hline
13               &                                 & 78.78               &                         \\ \hline
\end{tabular}
  \caption{\textit{Current Task Accuracy} for the different models for the Copy task distribution. The row with ``*" denotes the task at which the proposed model expanded. Capacity expansion technique allows our proposed model to clear more tasks than it would have cleared otherwise. The proposed  \textit{small-Lstm-Gem-Net2Net} model clears more levels than the \textit{large-Lstm-Gem} model. 
  }
  \label{tab::current-acc::copy}
\end{table}

\begin{table}[H]
\begin{tabular}{|c|c|c|c|}
\hline
\textbf{Task Id} & \textbf{small-Lstm-Gem-Net2Net} & \textbf{large-Lstm} & \textbf{large-Lstm-Gem} \\ \hline
1                & 75.99                           & 76.17               & 76.17                   \\ \hline
2                & 74.5 *                           & 76.57               & 75.63                   \\ \hline
3                & 75.6                            & 76.9                & 75.79                   \\ \hline
4                & 76.1                            & 76.03               & 75.49                   \\ \hline
5                & 75.29                           & 75.3                & 74.99                   \\ \hline
6                & 74.56                           & 74.9                & 74.38                   \\ \hline
7                &                                 & 75.38               &                         \\ \hline
8                &                                 & 75.5                &                         \\ \hline
9                &                                 & 75.38               &                         \\ \hline
10               &                                 & 75.07               &                         \\ \hline
11               &                                 & 74.8                &                         \\ \hline
\end{tabular}
  \caption{\textit{Current Task Accuracy} for the different models for the Associative Recall task distribution. The row with ``*" denotes the task at which the proposed model expanded. Capacity expansion technique allows our proposed model to clear more tasks than it would have cleared otherwise. The proposed  \textit{small-Lstm-Gem-Net2Net} model clears as many levels as the \textit{large-Lstm-Gem} model. 
  }
  \label{tab::current-acc::ar}
\end{table}

\begin{table}[H]
\begin{tabular}{|c|c|c|c|}
\hline
\textbf{Task Id} & \textbf{small-Lstm-Gem-Net2Net} & \textbf{large-Lstm} & \textbf{large-Lstm-Gem} \\ \hline
1                & 89.71                           & 90.59               & 90.59                   \\ \hline
2                & 77.67                           & 86.71               & 86.33                   \\ \hline
3                & 73.86                           & 88.08               & 86.51                   \\ \hline
4                & 74.437                          & 88.12               & 84.68                   \\ \hline
5                & 71.14                           & 89.32               & 79.13                   \\ \hline
6                & 67.84 *                         & 90.5                &                         \\ \hline
7                & 61.24                           & 90.97               &                         \\ \hline
8                &                                 & 90.3                &                         \\ \hline
9                &                                 & 89.89               &                         \\ \hline
10               &                                 & 88.49               &                         \\ \hline
11               &                                 & 81.64               &                         \\ \hline
  12               &                                 & 74.4                &\\ \hline
  13              &                                 & 67.8                &                         \\ \hline
\end{tabular}
  \caption{\textit{Current Task Accuracy} for the different models for the SSMNIST task distribution. The row with ``*" denotes the task at which the proposed model expanded. Capacity expansion technique allows our proposed model to clear more tasks than it would have cleared otherwise. The proposed  \textit{small-Lstm-Gem-Net2Net} model clears more levels than the \textit{large-Lstm-Gem} model. 
  }
  \label{tab::current-acc::ssmnist}
\end{table}

\subsubsection*{Previous Task Accuracy}

\begin{table}[H]
\begin{tabular}{|c|c|c|}
\hline
\textbf{large-Lstm} & \textbf{small-Lstm-Gem-Net2Net} & \textbf{large-Lstm-Gem} \\ \hline
58.15                           & 82.97               & 90.76                   \\ \hline
\end{tabular}
  \caption{\textit{{Previous} Task Accuracy} for the different models for the Copy task distribution. The proposed models are very close in performance to the \textit{large-Lstm-Gem} models and much better than the \textit{large-Lstm} models.}
  \label{tab::prev-acc::copy}
\end{table}

\begin{table}[H]
\begin{tabular}{|c|c|c|}
\hline
\textbf{large-Lstm} & \textbf{small-Lstm-Gem-Net2Net} & \textbf{large-Lstm-Gem} \\ \hline
75.2                           & 76.4             & 76.6                  \\ \hline
\end{tabular}
  \caption{\textit{{Previous} Task Accuracy} for the different models for the Associative Recall task distribution. The proposed models are very close in performance to the \textit{large-Lstm-Gem} models and much better than the \textit{large-Lstm} models.}
  \label{tab::prev-acc::ar}
\end{table}

\begin{table}[h]
\begin{tabular}{|c|c|c|}
\hline
\textbf{large-Lstm} & \textbf{small-Lstm-Gem-Net2Net} & \textbf{large-Lstm-Gem} \\ \hline
51.31                           & 83.01             & 82.92                 \\ \hline
\end{tabular}
  \caption{\textit{{Previous} Task Accuracy} for the different models for the SSMNIST task distribution. The proposed models are very close in performance to the \textit{large-Lstm-Gem} models and much better than the \textit{large-Lstm} models.}
  \label{tab::prev-acc::ssmnist}
\end{table}

\subsubsection*{Future Task Accuracy}

\begin{table}[H]
\begin{tabular}{|c|c|c|}
\hline
\textbf{large-Lstm} & \textbf{small-Lstm-Gem-Net2Net} & \textbf{large-Lstm-Gem} \\ \hline
56.55                           & 60.08               & 56.55                  \\ \hline
\end{tabular}
  \caption{\textit{{Future} Task Accuracy} for the different models for the Copy task distribution. }
  \label{tab::future-acc::copy}
\end{table}

\begin{table}[htbp]
\begin{tabular}{|c|c|c|}
\hline
\textbf{large-Lstm} & \textbf{small-Lstm-Gem-Net2Net} & \textbf{large-Lstm-Gem} \\ \hline
74.10                           & 73.22             & 73.14                  \\ \hline
\end{tabular}
  \caption{\textit{{Future} Task Accuracy} for the different models for the Associative Recall task distribution. The proposed models are very close in performance to the \textit{large-Lstm-Gem} models and much better than the \textit{large-Lstm} models.}
  \label{tab::future-acc::ar}
\end{table}

\begin{table}[htbp]
\begin{tabular}{|c|c|c|}
\hline
\textbf{large-Lstm} & \textbf{small-Lstm-Gem-Net2Net} & \textbf{large-Lstm-Gem} \\ \hline
37.01                           & 26.52            & 22.58                \\ \hline
\end{tabular}
  \caption{\textit{{Future} Task Accuracy} for the different models for the SSMNIST task distribution. The proposed models are very close in performance to the \textit{large-Lstm-Gem} models and much better than the \textit{large-Lstm} models.}
  \label{tab::future-acc::ssmnist}
\end{table}

\subsection{Accuracy of different models for different “task distributions” and tasks}

In the following figures, we plot the accuracy of the different models (\textit{small-Lstm-Gem-Net2Net}, \textit{large-Lstm-Gem} and \textit{large-Lstm} respectively) as they are trained and evaluated on different tasks for the Copy and the SSMNIST ``task distributions''. On the x-axis, we show the task on which the model is trained and on the y-axis, we show the accuracy corresponding to the different tasks on which the model is evaluated. We observe that for the \textit{large-Lstm} model, the high accuracy values are concentrated along the diagonal which indicates that the model does not perform well on the previous task. In the case of both \textit{small-Lstm-Gem-Net2Net} and \textit{large-Lstm-Gem} model, the high values are in the lower diagonal region indicating that the two models are quite resilient to catastrophic forgetting.

\begin{figure}
    \centering
    {\includegraphics[scale=0.6]{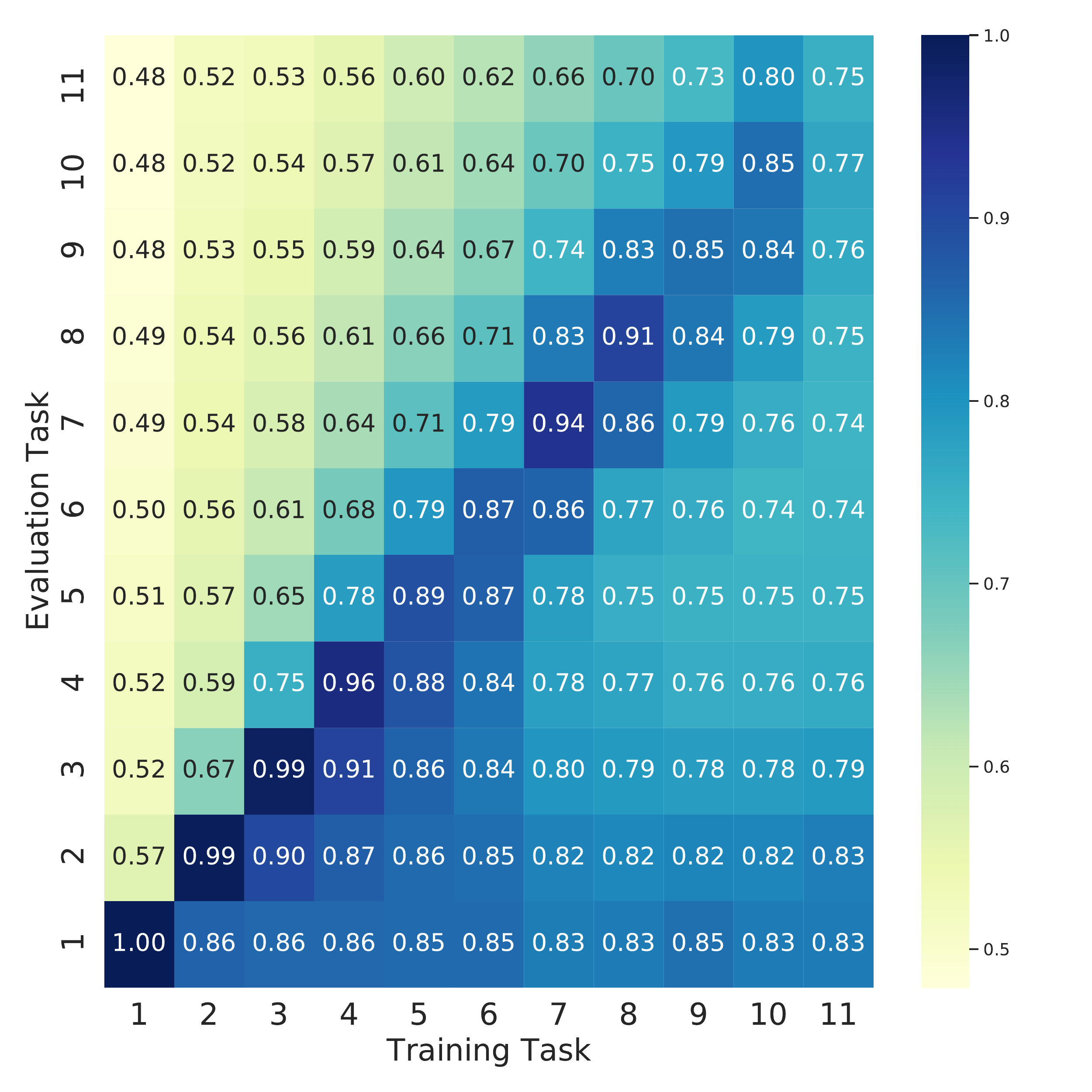}}
    \caption{\textit{small-Lstm-Gem-Net2Net} model for the Copy task}
\end{figure}

\begin{figure}
    \centering
    {\includegraphics[scale=0.6]{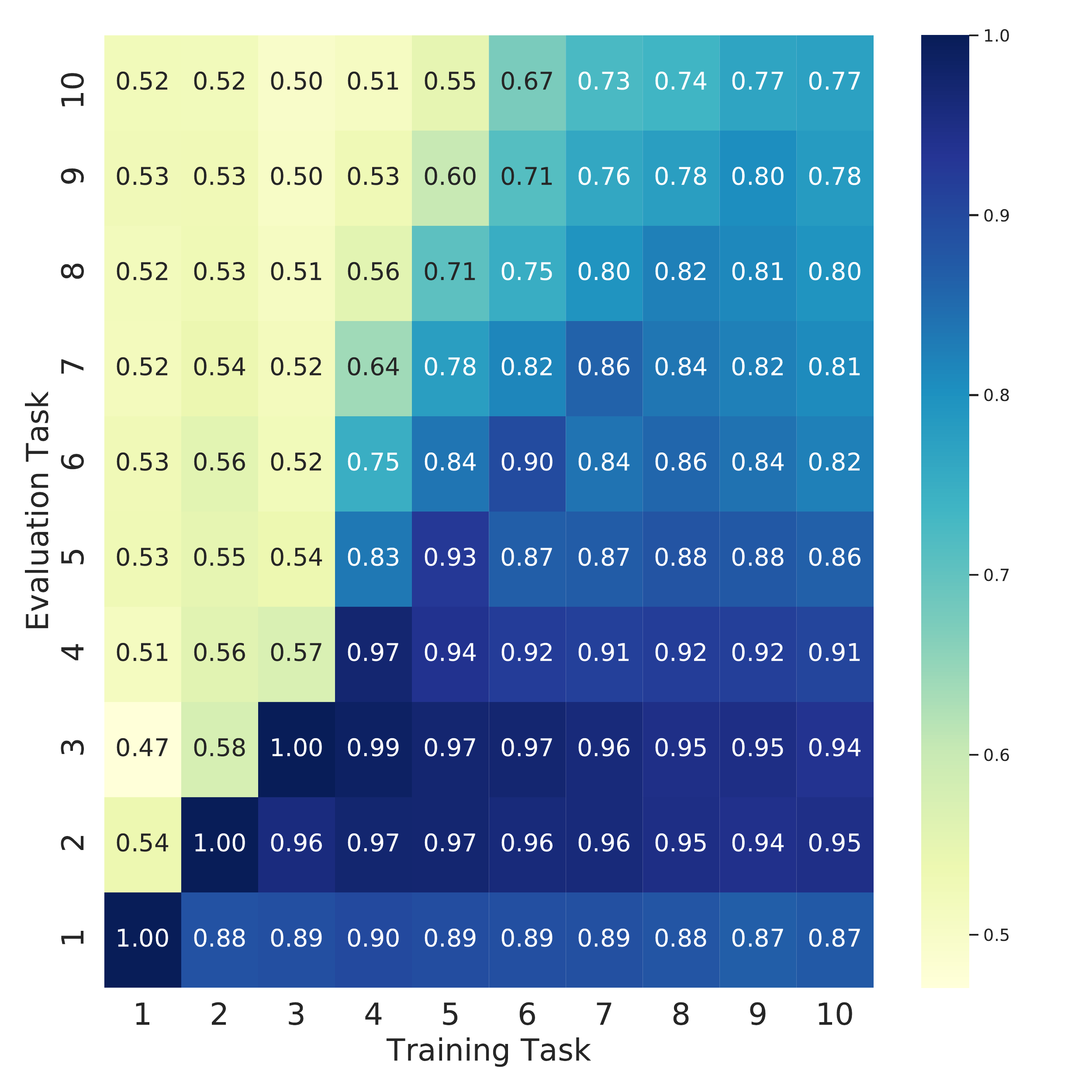}}
    \caption{\textit{large-Lstm-Gem} model for the Copy task}
\end{figure}

\begin{figure}
    \centering
    {\includegraphics[scale=0.7]{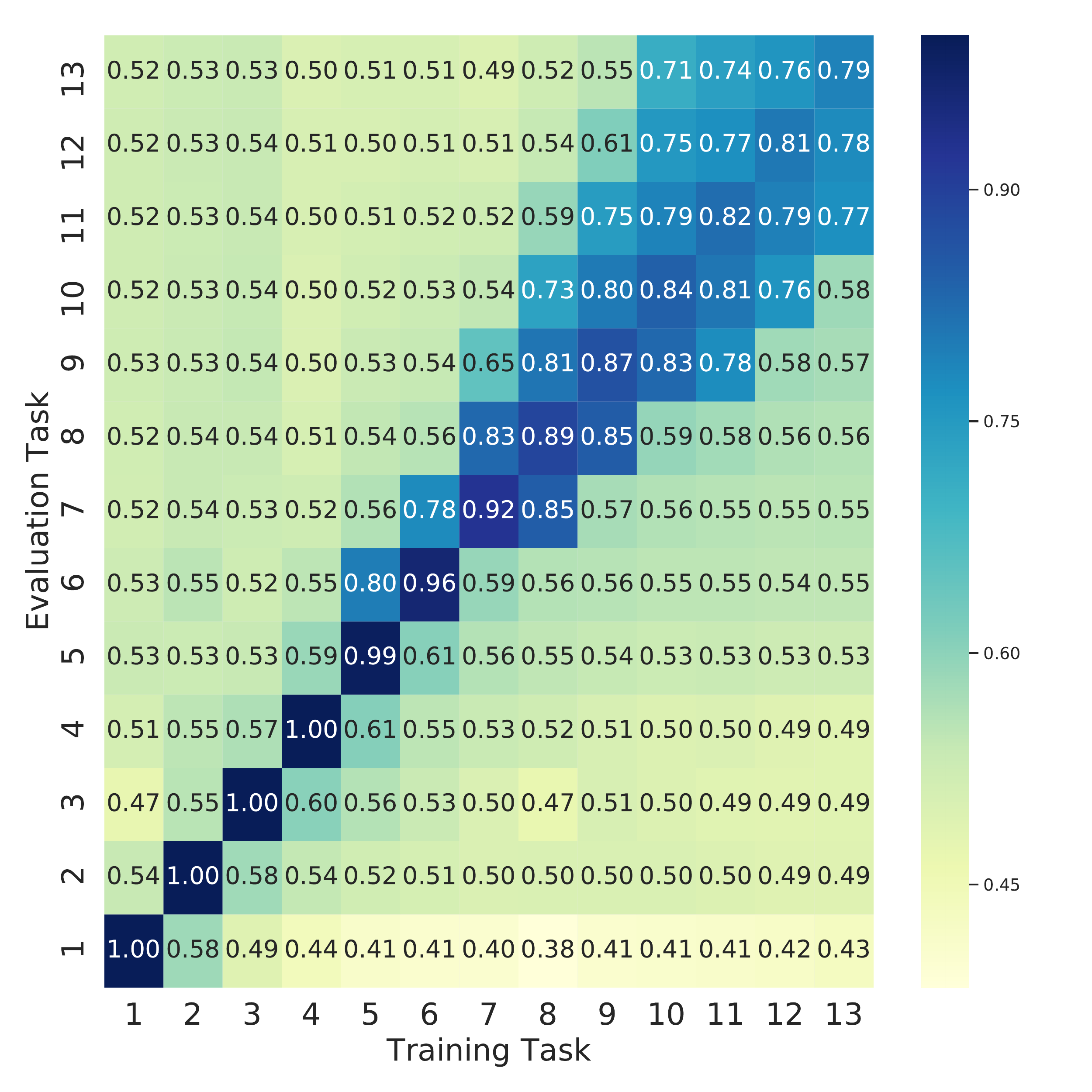}}
    \caption{\textit{large-Lstm} model for the Copy task}
\end{figure}

\begin{figure}
    \centering
    {\includegraphics[scale=0.6]{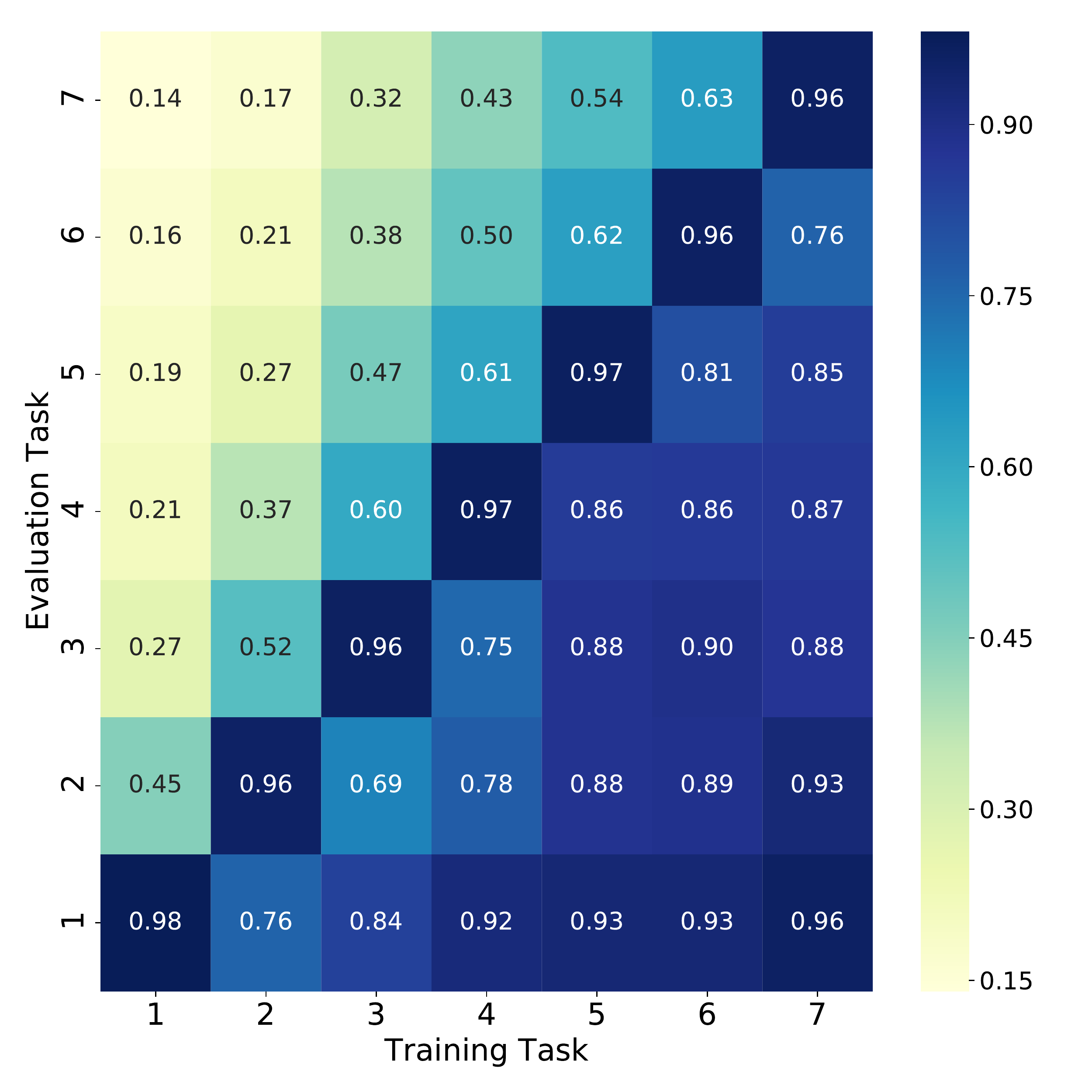}}
    \caption{\textit{small-Lstm-Gem-Net2Net} model for the SSMNIST task}
\end{figure}

\begin{figure}
    \centering
    {\includegraphics[scale=0.6]{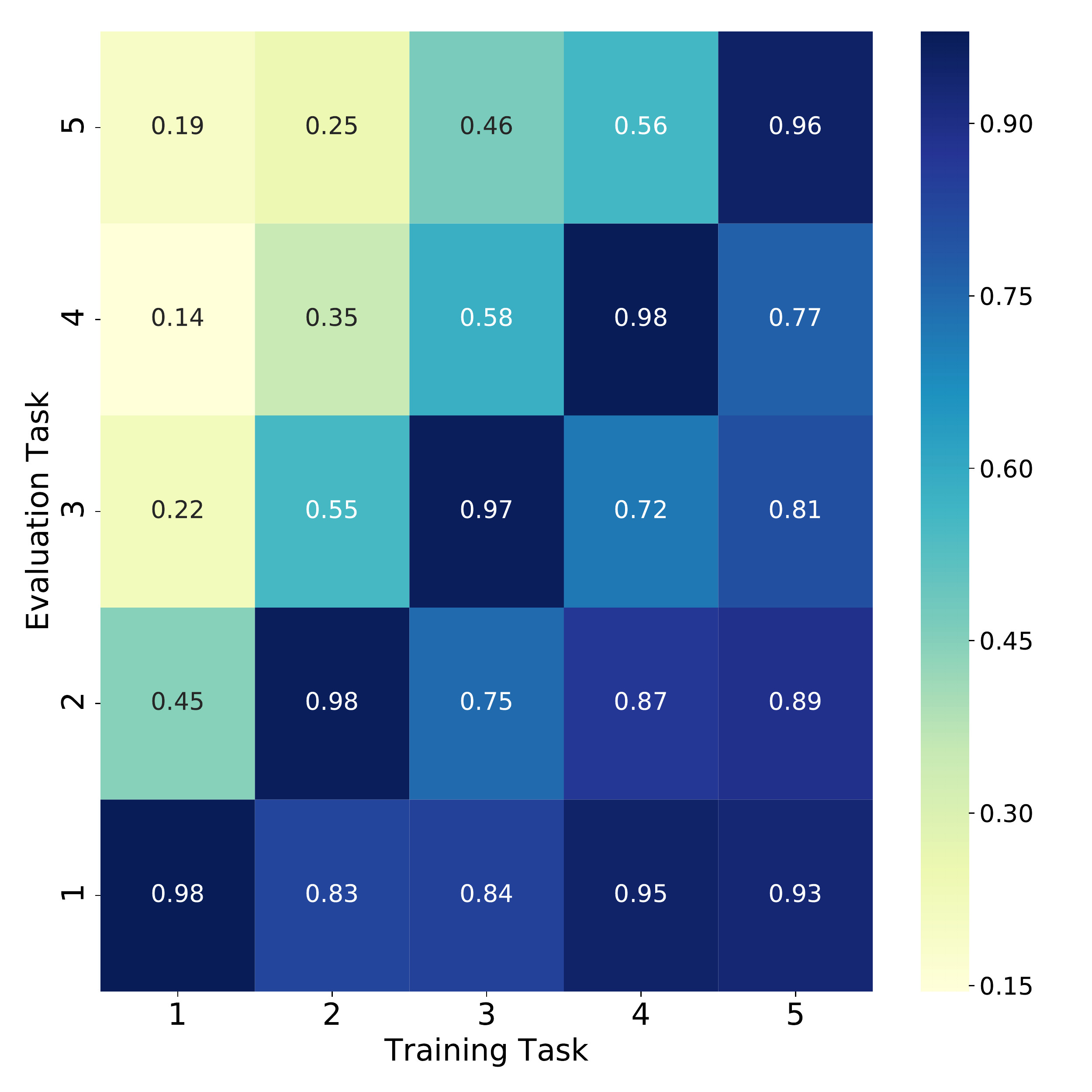}}
    \caption{\textit{large-Lstm-Gem} model for the SSMNIST task}
\end{figure}

\begin{figure}
    \centering
    {\includegraphics[scale=0.7]{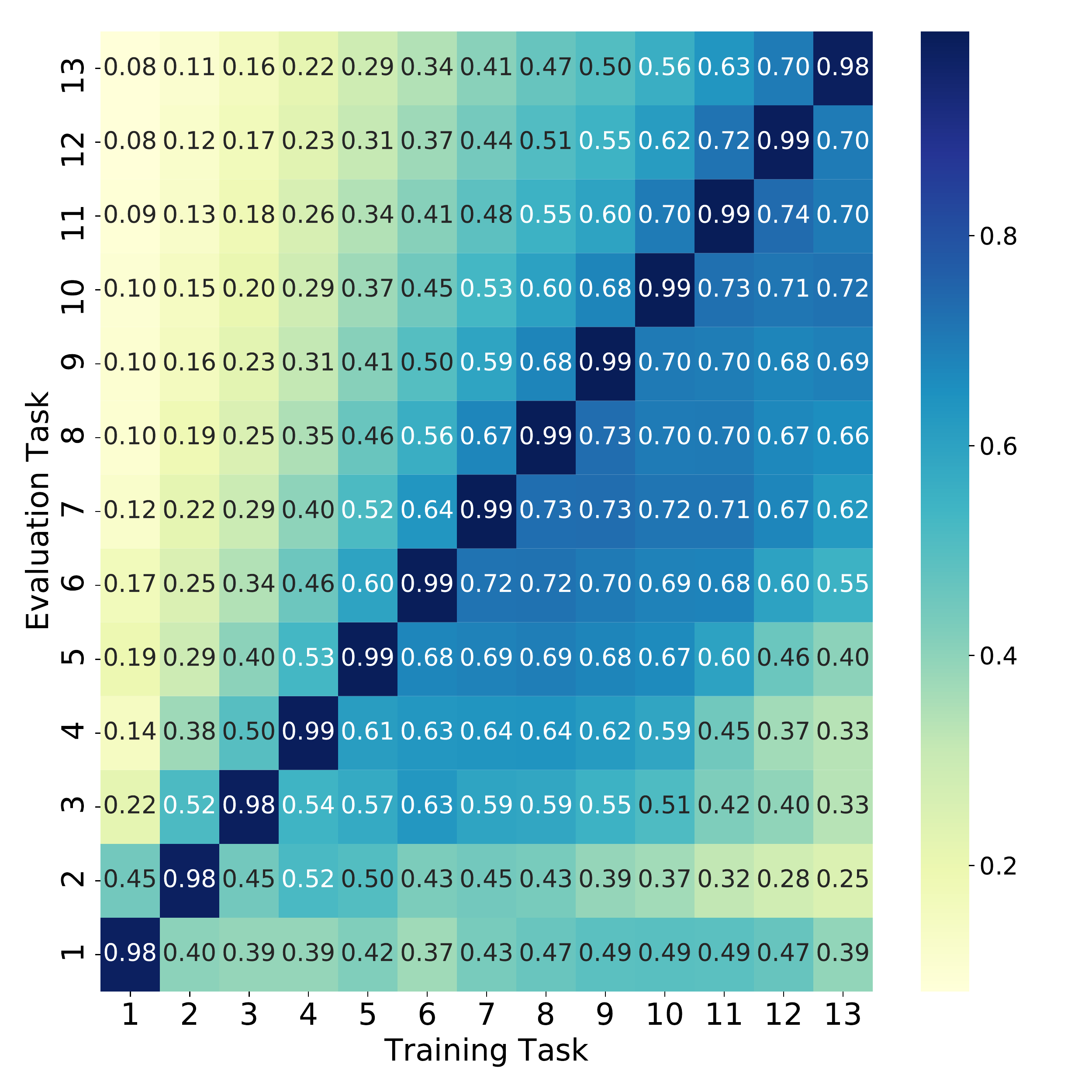}}
    \caption{\textit{large-Lstm} model for the SSMNIST task}
\end{figure}

\end{document}